\documentclass{article}
\usepackage[dvipsnames, table]{xcolor}
\usepackage{amssymb}
\usepackage{amsmath}
\usepackage{amsthm}
\usepackage[a4paper]{geometry}
\usepackage[round]{natbib}
    
\usepackage{graphicx}
\usepackage{booktabs}
\usepackage{tikz} 
\usepackage{float}
\usepackage[accsupp]{axessibility}  
\usepackage{multirow}
\usepackage{amsfonts}
\usepackage{comment}

\usepackage{caption}
\usepackage{subcaption}
\usepackage{booktabs}       
\usepackage[utf8]{inputenc} 
\usepackage[T1]{fontenc}    
\usepackage{url}            
\usepackage{amsfonts}       
\usepackage{nicefrac}       
\usepackage{microtype}      
\usepackage{xcolor}         
\usepackage{subfiles}
\usepackage{enumitem}
\usepackage{multicol}
\usepackage{multirow}
\usepackage{arydshln}
\usepackage{comment}
\usepackage{listings}
\usepackage{array}
\usepackage{adjustbox}
\usepackage{wrapfig}
\usepackage[colorlinks,linktoc=true,
             citecolor= Cerulean!85!green!90!black,
             linkcolor=YellowOrange,
             urlcolor=RubineRed!85!black,
             ]{hyperref}

\usepackage[ruled,linesnumbered]{algorithm2e}
\SetKwData{KwInput}{Input:}

\SetKwFor{ForEach}{for each}{do}{end for}
\DontPrintSemicolon

\definecolor{darkgreen}{RGB}{126,171,85}
\definecolor{darkblue}{RGB}{47,110,186}
\definecolor{darkred}{RGB}{192,0,0}

\setlist[itemize]{noitemsep,leftmargin=*,topsep=0.3ex}

\usepackage{framed}

\usepackage{amsmath}
\usepackage{amssymb}
\usepackage{mathtools}
\usepackage{amsthm}

\usepackage[capitalize,noabbrev]{cleveref}
\theoremstyle{plain}
\newtheorem{theorem}{Theorem}
\newtheorem{proposition}[theorem]{Proposition}

\theoremstyle{definition}

\theoremstyle{remark}
\newtheorem{remark}{Remark}
\usepackage[textsize=tiny]{todonotes}

\usepackage{lipsum}

\title{Approximate Nullspace Augmented Finetuning for Robust Vision Transformers}

\author{
Haoyang Liu\textsuperscript{1}, Aditya Singh\textsuperscript{2}, Yijiang Li\textsuperscript{3}, Haohan Wang\textsuperscript{1} \\
\\
\textsuperscript{1}University of Illinois Urbana-Champaign\\
\textsuperscript{2}Sorted Technologies\\
\textsuperscript{3}University of California, San Diego\\
\textsuperscript{1}\texttt{\{hl57, haohanw\}@illinois.edu} \\
\textsuperscript{2}\texttt{aditya.singh@sortedtech.io}\\
\textsuperscript{3}\texttt{yijiangli@ucsd.edu}\\
}
\date{}

\begin{document}
\maketitle

\newcolumntype{C}[1]{>{\centering\arraybackslash}p{#1}}

\begin{abstract}
Enhancing the robustness of deep learning models, particularly in the realm of vision transformers (ViTs), is crucial for their real-world deployment. In this work, we provide a finetuning approach to enhance the robustness of vision transformers inspired by the concept of nullspace from linear algebra. Our investigation centers on whether a vision transformer can exhibit resilience to input variations akin to the nullspace property in linear mappings, which would imply that perturbations sampled from this nullspace do not influence the model's output when added to the input. We start from the observation that many existing ViTs satisfy this property because their patch embedding layer has a non-trivial nullspace. Then, we extend the notion of nullspace to nonlinear settings and demonstrate that it is possible to synthesize approximate nullspace elements for ViT's encoder blocks through optimization. Finally, we propose a finetuning strategy for ViTs wherein we augment the training data with synthesized approximate nullspace noise. We find that our finetuning approach significantly improves the models' robustness to both adversarial and natural image perturbations.\footnote{Code is available at: \url{https://github.com/Liu-Hy/ns-vit}.}
\end{abstract}

\section{Introduction}
\label{sec:introduction}

The field of computer vision has experienced significant advances, marked by the emergence of Vision Transformers (ViTs)~\citep{dosovitskiy2020vit} as a notable milestone. Following this advancement, a series of architectural refinements have been explored~\citep{ali2021xcit, li2022contextual, liu2021Swin}, paving the way for the development of vision foundation models~\citep{kirillov2023segment, zou2023segment} through the scaling up of both the model and dataset. Despite these strides, robustness continues to be a pivotal concern for their practical deployment, as they exhibit fragility in the face of imperceptible (adversarial) and perceptible perturbations.

Adversarial samples are generated by adding imperceptible noises to the input, aiming to cause the model to produce incorrect and overly confident predictions~\citep{adv_ex_0, adv_ex_1, adv_ex_2}. Perceptible perturbations are artifacts that arise from various operations, such as JPEG compression, simulated weather effects (fog, snow), or adjustments to the image's brightness, hue, or contrast, to name a few~\citep{hendrycks2018benchmarking}. The semantic content of the image however, remains unchanged after perceptible or imperceptible perturbations. Hence, we expect the model to output similar predictions for perturbed and unperturbed images.

Applying transformations to the input during training, known as data augmentation, is one of the widely employed techniques for improving robustness. The underlying goal of applying augmentations is to enforce invariance (i.e., consistency) under a predefined set of perturbations. To induce adversarial robustness, worst-case adversarial perturbations are first identified through an optimization procedure and then used to train the model~\citep{madry2018towards, trades}. For robustness against perceptible noise, augmentation strategies have evolved from simple transformations such as horizontal flips and rotations to more complex augmentations like MixUp~\citep{zhang2018mixup}, CutMix~\citep{yun2019cutmix},  and AugMix~\citep{hendrycks_2020augmix}.  

There is an observable divide in the treatment of these two types of robustness~\citep{liu2023towards}. Adversarial noises are generated via an optimization process, whereas augmentations are defined heuristically often by domain experts. It has also been observed that standard data augmentation strategies, in isolation, do not improve adversarial robustness~\citep{gowal2021uncovering, pmlr-v119-rice20a, rebuffi_aug}. Additionally, adversarial training (training with adversarial perturbations) often leads to a drop in performance on non-adversarial images~\citep{madry2018towards, trades, clarysse2023why}.

\begin{figure}
    \centering
    \includegraphics[width=0.7\textwidth]{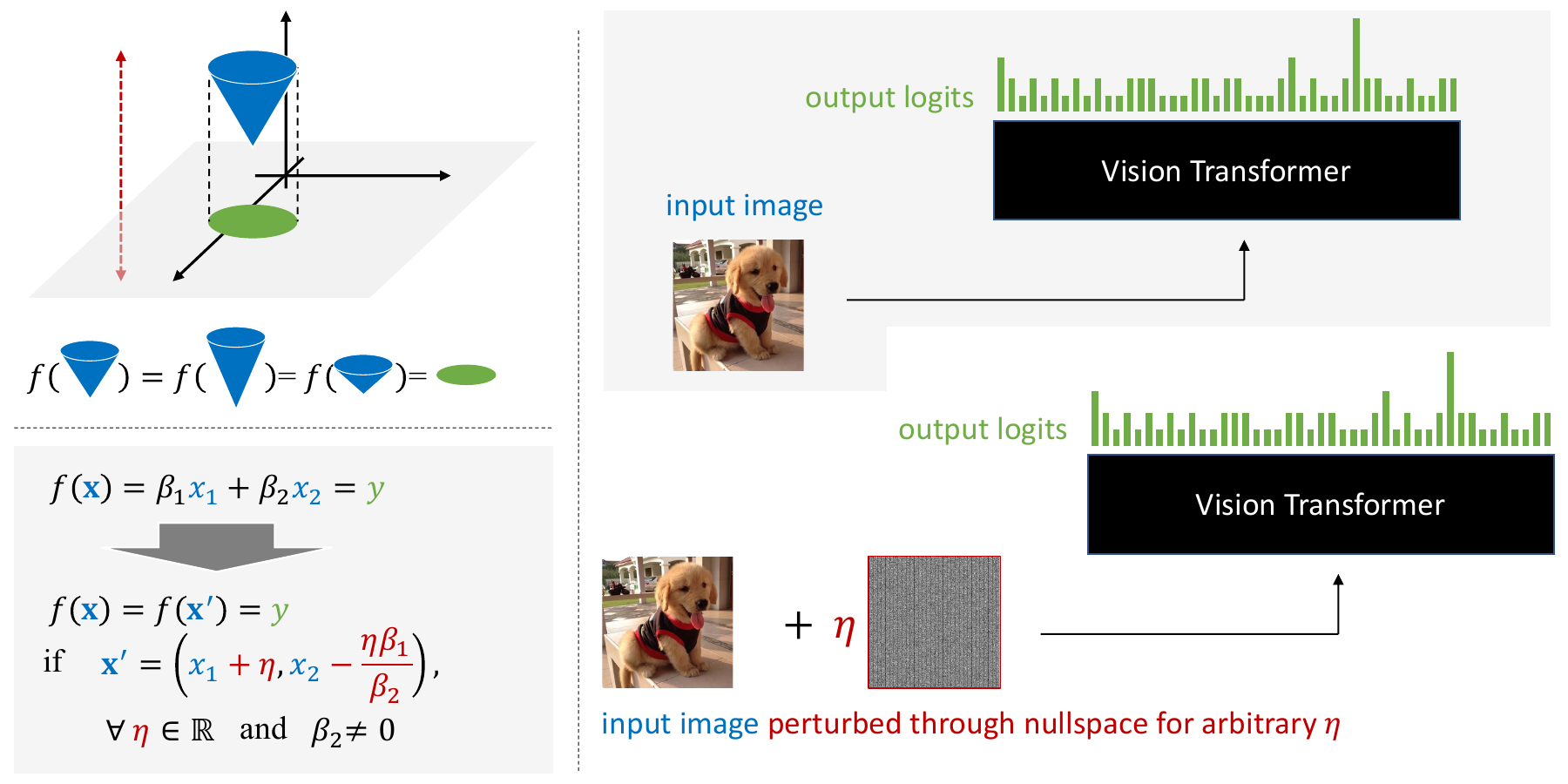}
    \caption{\textbf{An illustration of the nullspace in three cases (projection function, left top; linear function, left bottom; vision transformer, right)}. 
    For these three cases, there exists some {\color{darkred} nullspace}, such that the {\color{darkgreen} output} of the function with respect to the {\color{darkblue} input} will remain unperturbed regardless of the perturbation strength. Also, the {\color{darkred} nullspace} is function-specific (model-specific) and will not vary for different samples. 
    }
    \label{fig:main}
\end{figure}

Here, we consider robustness a property of the model and thus agnostic to the noise type. To this end, we consider the nullspace as the central theme of our study. The nullspace, a fundamental concept in linear algebra, refers to the subspace of a domain that is mapped to zero by a linear mapping. By definition, any vector from the nullspace, when added to the input of the linear mapping, does not alter its output. In \cref{fig:main}, we present the concept of nullspace from different perspectives.

This paper first identifies that most off-the-shelf pre-trained ViT models exhibit a nontrivial nullspace due to the linear patch embedding layer. Since this layer is the first block of a ViT, any invariance to it implies invariance to the entire model. Consequently, a nontrivial nullspace also exists for ViTs. To further explore robustness, we define the approximate nullspace of the transformer encoder and use optimization methods to synthesize noise vectors approximating nullspace properties for nonlinear blocks. Finally, we propose fine-tuning the model using these synthesized nullspace-like elements as additive training data augmentation. This approach enlarges the approximate nullspace, enhancing the model's robustness. The main contributions of our paper include:

\begin{itemize}
\item We demonstrate connections between the robustness of vision transformers to the algebraic notion of nullspace, substantiated by experimental results showing that enlarging the approximate nullspace effectively improves the model robustness.
\vspace{0.3em}
\item We conduct comprehensive analysis on the existence of nullspace within transformer models. We establish the existence of nullspace at the patch embedding layer, and empirically identify an approximate nullspace at the nonlinear encoder level of transformers by validating their algebraic properties.
\vspace{0.3em}
\item We propose an effective data augmentation method by exploiting and enlarging the model's approximate nullspace, which enhances model robustness without architectural modifications and only involves fine-tuning with minimal additional data. Our method is empirically validated across multiple benchmark datasets, showing significant robustness improvements against adversarial and out-of-distribution scenarios. 

\end{itemize}



\section{Related Work}
\label{sec:related_work}
\textbf{Data augmentation and Invariance:}
Data augmentation enforces invariance by training models to predict consistently across different input views, offering a theoretical improvement in estimating statistical risk~\citep{chen2020group,shao2022theory}. However, incorrect augmentation choices can degrade performance~\citep{chen2020group, lyle2020benefits, shao2022theory}. Early image augmentations such as flipping, cropping and rotation have evolved into advanced techniques such as MixUp~\citep{zhang2018mixup}, CutMix~\citep{yun2019cutmix}, and strategies for chain augmentations, including AutoAugment for policy optimization, TrivialAugment~\citep{Muller_2021_ICCV}, and RandAug~\citep{randaug}. AugMix~\citep{hendrycks_2020augmix} combines transformations with a consistency loss, while differentiable augmentations optimize transformations for specific tasks~\citep{li2020dada, chatzipantazis2023learning}. \citet{inv_learn} frame data augmentation as an invariance-constrained learning problem, using a relaxed invariance notion to model augmentation distributions. Unlike these approaches, our work avoids reliance on pre-defined augmentations.

\textbf{Robustness in ViTs:}
Research highlights Vision Transformers (ViTs) as more robust than Convolutional Neural Networks (CNNs)\citep{shao2021adversarial, paul2021vision}, with adversarial examples that exhibit low transferability between these architectures\citep{mahmood2021robustness}, although some studies offer counterpoints~\citep{NEURIPS2021_bai}. ViTs demonstrate insensitivity to patch-based transformations that distort semantics, relying on robust but nonindicative features~\citep{PatchRobust_22}. Robustness-enhancing methods for transformer-based models are often model-agnostic, using data augmentation~\citep{xiao2023masked, esser2021taming, steiner2022how, liu2022tokenmix} and regularization~\citep{chen2022when, steiner2022how, chefer2022optimizing}, consistent with broader robustness frameworks~\citep{wang2022generalizing, liu2023trustworthy}. For example, \citet{xiao2023masked} masks image patches using class activation maps and refills them with random samples, while \citet{chen2022when} adopts sharpness-aware optimization for a smoother loss landscape. However, these approaches focus on external modifications or optimization, often neglecting the intrinsic properties of the model.

\textbf{Nullspace and Neural Networks:}
The study of nullspaces in neural networks began with \citet{mlp_null}, who explored MLPs' universal approximation by comparing input nullspaces and outputs. Using the \emph{learning XOR} example, they demonstrated that hidden layers enable MLPs to map inputs to targets even if the targets reside in the nullspace of the inputs. More recently, \citet{ghosts} mathematically analyzed nullspaces in fully connected networks.

In applications, \citet{Wang_2021_CVPR} leveraged nullspaces in continual learning to map new tasks to the nullspace of existing ones. As a novel architecture, NullSpaceNet~\citep{nullnet} mapped inputs from the same category to a joint nullspace rather than a feature space.

\section{Nullspace and Invariance}
\label{sec:nullspace}
When a mapping $f: \mathcal{X}\rightarrow\mathcal{Y}$ is invariant to some additive noise $\mathbf{v}$, it implies the following: 
\begin{equation}
\label{eq:invariance}
    f(\mathbf{x}+\mathbf{v}) = f(\mathbf{x}) \quad  \forall \mathbf{x} \in \mathcal{X}.
\end{equation}
This invariance has interesting connections to the concept of \emph{nullspace} in linear algebra. Formally, the nullspace of a linear mapping $f$ is a set $\mathcal{N}$ identified by $\mathcal{N} = \{\mathbf{v} \in \mathcal{X} | f(\mathbf{v}) = 0\}$. 
For a non-trivial nullspace $\mathcal{N} \neq \phi$, we have $f(\mathbf{x}+\mathbf{v}) = f(\mathbf{x}), \forall \mathbf{v} \in \mathcal{N}, \forall \mathbf{x} \in \mathcal{X}$. We can interpret this by saying that the linear mapping is invariant to the noise vector sampled from its nullspace. For brevity, we refer to this noise vector as \textbf{nullspace noise}.


\subsection{Non-trivial nullspace}
\label{sec:vits}
Vision transformer \cite{dosovitskiy2020vit} is a function $f_\omega$ with $\omega$ as the trainable weights. The function takes as input an image $\mathbf{x} \in \mathcal{X}^{c\times h\times w}$ and outputs a classification response $\mathbf{y} \in \mathcal{Y}^k$ over $k$ categories. $c$ is the number of channels (typically $3$ for red, green, and blue), $h$, $w$ correspond to height and width of the input image. This neural network function can be broken down into $3$ stages, namely:
\begin{itemize}[leftmargin=*]
    \item \textit{patch embedding stage}, $f_\theta: \mathcal{X}^{c\times r \times r} \rightarrow \mathcal{U}^{d}$. This steps projects the input image patch of predetermined dimensions $c$, $r$ and $r$ to a one-dimensional embedding of length $d$. It is ensured that patches have no overlaps between them. Hence, the number of such non-overlapping patches generated from the input image are $m = \frac{h\times w}{r^2}$.
    \item \textit{self-attention stage}, $f_\phi:\mathcal{U}^{(m+1)\times d} \rightarrow \mathcal{V}^{(m+1)\times d}$. In the next step, the generated patch embeddings are passed through layers of self-attention modules to process long range interactions amongst them. Apart from the $m$ patch embeddings an additional embedding in form of \texttt{cls} token is utilised in this step.
    \item \textit{classification stage}, $f_\psi: \mathcal{V}^{d} \rightarrow \mathcal{Y}^{k}$. The final step is to perform the $k$-way classification. For this, we simply keep the processed encoding corresponding to \texttt{cls} token and project it through a linear classification layer.
\end{itemize}
\begin{wrapfigure}{r}{0.5\textwidth}
\vspace{-4.7mm}
\centering
\small
\captionof{table}{\textbf{Nullspace dimensions for pre-trained ViT models.} Nullspace is trivial ($\mathbf{0}$) when embedding dimension exceeds input dimension.}
\begin{tabular}{lccr}
\toprule
Model & Patch Size & Emb. Dim. & Null Dim. \\
\midrule
\texttt{tiny}   & $16 \times 16$ & 192  & 576  \\
\texttt{small}  & $32 \times 32$ & 384  & 2688 \\
                & $16 \times 16$ & 384  & 384  \\
\texttt{base}   & $32 \times 32$ & 768  & 2304 \\
                & $16 \times 16$ & 768  & 2    \\
                & $8 \times 8$   & 768  & 0    \\
\texttt{large}  & $32 \times 32$ & 1024 & 2048 \\
                & $16 \times 16$ & 1024 & 0    \\
\bottomrule
\end{tabular}
\label{tab:empirical_nullspace}
\end{wrapfigure}

Since the first layer of the ViT is a linear mapping, according to the rank-nullity theorem, it always has a non-trivial nullspace if $cr^2 > d$. In practice, for many ViT-based architectures, we find that this is the case. In \cref{tab:empirical_nullspace}, we report the identified nullspace dimensions for off-the-shelf pre-trained ViT models.

\begin{figure}[t!]
  \centering
    \begin{subfigure}[c]{0.3\columnwidth}
    \includegraphics[width=1\linewidth]{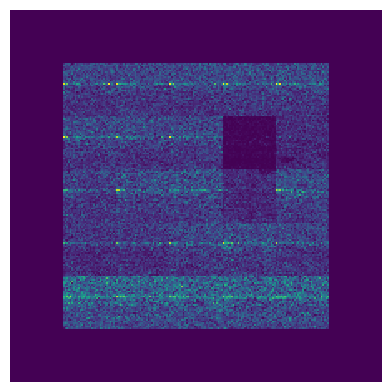}
    \caption{\footnotesize Sample null-space noise}
  \end{subfigure}%
  \hfill
    \begin{subfigure}[c]{0.3\columnwidth}
    \includegraphics[width=1\linewidth]{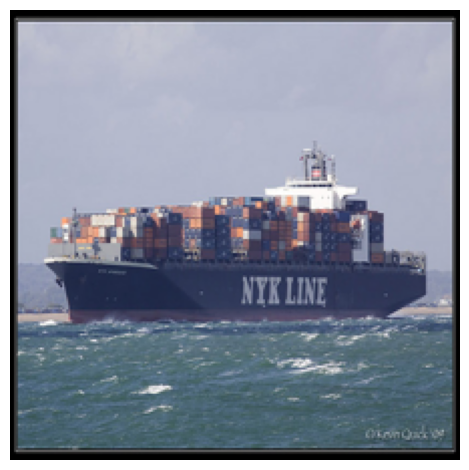}
    \caption{\footnotesize Clean input image}
  \end{subfigure}%
  \hfill
  \begin{subfigure}[c]{0.3\columnwidth}
    \includegraphics[width=1\linewidth]{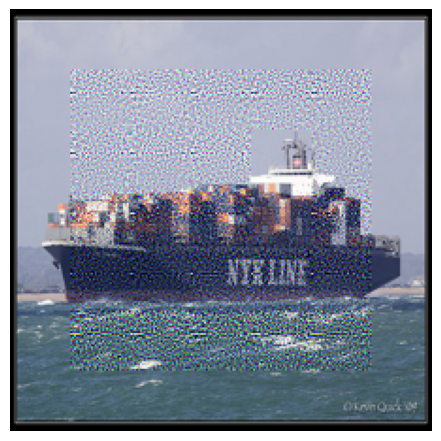}
    \caption{\footnotesize Noisy input image}
  \end{subfigure}%

  \caption{\textbf{An example of nullspace noise.} We show (a) sample input image, (b) noise generated by the basis vectors of the nullspace and (c) noisy image as a result of adding the nullspace noise to the input. Model's predictions for the clean and noisy inputs are identical. } 
  \label{fig:example_ns}
\end{figure}

Given the weights of the patch embedding layer $f_\theta$, finding its nullspace is a standard practice \citep{kwak04, strang09, Strang2009Introduction}.
Let $B_\theta = \{\mathbf{b}_1, \mathbf{b}_2, \dots \mathbf{b}_k\}$ be the $k$ basis vectors for this nullspace, we can sample an element from $\mathcal{N}_\theta$ as:
\begin{equation}
    \mathbf{v} = \lambda_1\mathbf{b}_1 + \lambda_2\mathbf{b}_2 + \dots + \lambda_d\mathbf{b}_k.
\label{nu_theta}
\end{equation}
The property of such a sample will be that the output of the patch embedding will effectively remain preserved, $f_{\theta}(\mathbf{x}+\mathbf{v}) = f_{\theta}(\mathbf{x})$. Since the output after the first layer remains unaffected,
the final output of the classification remains unchanged. In \cref{fig:example_ns}, we provide visualization of noise synthesized using basis vectors. This noise can be added to \textit{any} input image with complete invariance. In \cref{sec:vits:2}, we explore if it possible to learn a nullspace-like counterpart for the non-linear blocks of ViTs.

\subsection{The Encoder-level nullspace}
\label{sec:vits:2}
So far we have demonstrated that a non-trivial nullspace exists for the patch embedding layer, and hence the entire vision transformer is invariant to all perturbations in that space. We move further down the structure of ViT and investigate whether the encoder is also invariant to certain perturbations. The self-attention layer is non-linear, which means the notion of nullspace cannot be directly applied to $f_\phi$. However, the \emph{invariance} property that can be implied from the nullspace of linear functions, that any vector from this set will not alter the function's output when added to any input, is still desirable in the nonlinear case when it comes to the robustness of neural models. In fact, data augmentation can often be formulated as a process of adding noise to the input and enforcing invariance. Therefore, to study the ViTs' inherent invariance to input perturbations, we extend the notion of nullspace to the nonlinear setting and define the \emph{Generalized Nullspace}, $\tilde{\mathcal{N}}_\phi$, of the transformer encoder $f_\phi$, as below:
\begin{equation}
\label{eq: invariance1}
   \tilde{\mathcal{N}}_\phi = \{\mathbf{v} | f_\phi(\mathbf{u}+\mathbf{v})=f(\mathbf{u}) \quad \forall \mathbf{u} \in \mathcal{U}\},
\end{equation}
Here, we use the tilde accent $\tilde{\cdot}$ to distinguish $\tilde{\mathcal{N}}_\phi$ from the conventional nullspace $\mathcal{N}_\phi$. We term it the Generalized Nullspace because it depicts invariance in both linear and nonlinear settings, and that for a linear fuction $f_\theta$ we have  $\mathcal{N}_\theta \subseteq \tilde{\mathcal{N}}_\theta$, since any vector sampled from the conventional nullspace of a linear function satisfies this invariance property.
If such a set exists, it directly implies that the transformer model is robust to certain perturbations in the input space. Our theoretical analysis established the following sufficient conditions for the existence of a nontrivial generalized nullspace. (The complete proof is given in Appendix A.)
\begin{proposition}
Consider a self-attention layer with $h$ heads and $\{(\mathbf{Q}_i, \mathbf{K}_i, \mathbf{V}_i)\}_{i=1}^h$ as its query, key and value projection matrices. If the following conditions are met 
\begin{enumerate}[itemsep=-2pt, topsep=-2pt]
    \item \(\mathbf{Q}_i\mathbf{K}_i^\top\) is symmetric for \(i = 1, \dots, h\)
    \item The row space \(\operatorname{R}(\mathbf{V}_i^\top) \subseteq \operatorname{R}(\mathbf{Q}_i\mathbf{K}_i^\top)\) for \(i = 1, \dots, h\) 
    \item for some \(m \neq n\), \(\mathbf{Q}_m\mathbf{K}_m^\top\) has colinearity with \(\mathbf{Q}_n\mathbf{K}_n^\top\), i.e. for some \(k\) the \(k\)th row of \(\mathbf{Q}_m\mathbf{K}_m^\top\), denoted as $\mathbf{r}_{m, k}$, satisfies \(\mathbf{r}_{m, k} \neq \mathbf{0} \) and \(\mathbf{r}_{m, k} \in \operatorname{R}(\mathbf{Q}_n\mathbf{K}_n^\top)\)
\end{enumerate}

then there exists at least one $\mathbf{W}$ such that
$\mathbf{W} \neq \mathbf{0}$ and
$\operatorname{head}_i(\mathbf{X+W})  = \operatorname{head}_i(\mathbf{X})$
for all attention head $i$ in this layer and arbitrary $\mathbf{X}$.
\end{proposition}

\begin{remark}
Condition 1 can be met if $\mathbf{Q}_i$ and $\mathbf{K}_i$ satisfy some special relation. For example, let $\mathbf{PDP}^{-1}$ be a diagonalization of a real symmetric matrix $\mathbf{A}$. If $\mathbf{Q}_i = \mathbf{BP}$ and $\mathbf{K}_i = \mathbf{B}(\mathbf{P}^{-1})^\top \mathbf{D}$, then we have $\mathbf{Q}_i\mathbf{K}_i^\top = \mathbf{BAB}^\top$ to be symmetric.

In addition, evidence has shown that,  
$\mathbf{Q}_i\mathbf{K}_i^\top$ can be empirically symmetric, especially for ViTs,
when the attention heads 
are visualized and correlation of parameters 
is calculated \citep{yeh2023attentionviz}.
\end{remark}

\begin{figure}[h!]
  \centering
  \begin{subfigure}[c]{0.46\columnwidth}
  \centering
    \includegraphics[width=1\linewidth]{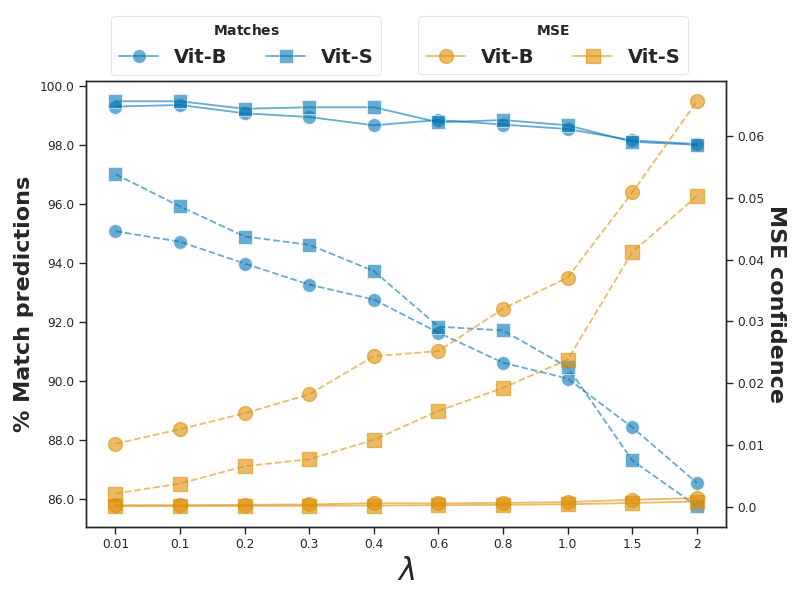}
    \caption{\small{Noise influence on the model output under different regularization strengths}}
  \end{subfigure}%
  \hfill 
  \begin{subfigure}[c]{0.52\columnwidth}
  \vspace*{5mm}
  \centering
    \includegraphics[width=0.8\linewidth]{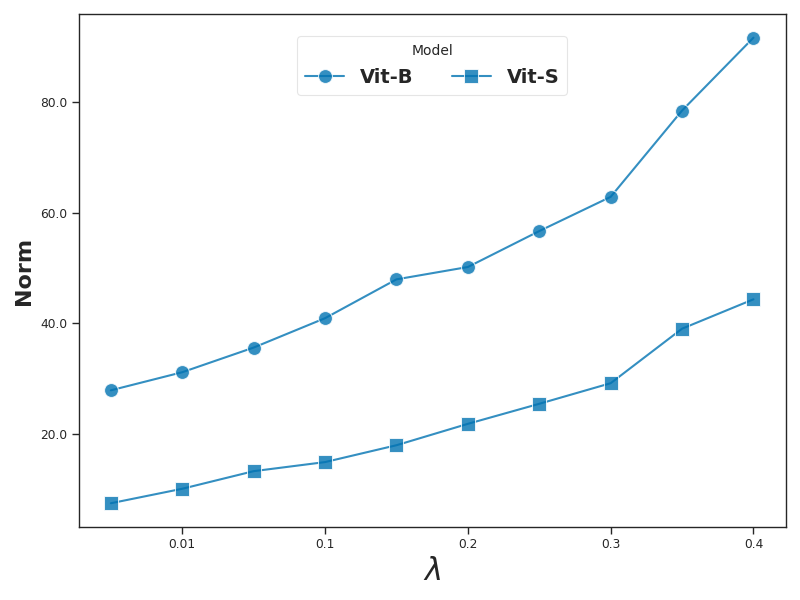}
    \caption{\small{$\ell_2$ norm of learned noise under different regularization strengths}}
  \end{subfigure}%
  \caption{\textbf{Exploratory experiments on the generalized nullspace.} (a) Solid lines (--) represents the model performance under the learned noise, and dashed lines ($\cdot\cdot\cdot$) represent the performance after random permutation of the elements of the learned noise vector. (b) by changing the regularization strengths, we explore noise in the generalized nullspace at different magnitudes.}
  \label{fig:enc_ns}
\end{figure}

\subsection{Synthesizing (approximate) nullspace noise}

Although our theory suggests a sufficient condition for the existence of generalized nullspace, analytically finding $\tilde{\mathcal{N}}_\phi$ or probing its existence for generic transformers is challenging. Thus, as an exploratory experiment, we employ a numeric method: we search for individual element, $\tilde{\mathbf{v}}_\phi$, of this set. This element is an additive perturbation that brings minimal influence to the output of $f_\phi$ on the data distribution. We introduce a regularization term on the norm of $\tilde{\mathbf{v}}_\phi$ to prevent the trivial solution of $\mathbf{0}$.
\begin{equation}
  \mathcal{L_\phi}(\tilde{\mathbf{v}}) = \underbrace{\mathbb{E}_{\mathbf{u} \in \mathcal{D}} \,
  \lVert f_\psi(f_\phi^{0}(\mathbf{u}+\tilde{\mathbf{v}}))
  - f_\psi(f_\phi^{0}(\mathbf{u}))\rVert}_{\mathfrak{I}}\,
  - \lambda \log(\lVert \tilde{\mathbf{v}} \rVert).
\label{n_opt} 
\end{equation}

Here, $\lVert \cdot \rVert$ is the $\ell_2$ norm, $f_\phi^0$ is the representation of the \texttt{cls} token output by $f_\phi$, and $\lambda$ is the regularization coefficient. $\mathfrak{I}$ resembles a weaker notion of invariance compared to \cref{eq:invariance}. 

\cref{n_opt} minimizes the $\ell_2$ norm between the predicted logits with and without the noise. 
Alongside the self-attention stage, we have also incorporated the classification stage into the loss, since the target of our study is to minimize the impact on the final output of the network. To learn the noise vector, we initialize $\tilde{\mathbf{v}}$ by sampling from a uniform distribution, and minimize the loss with gradient descent. We use ViT-S and ViT-B models with patch size 32 for evaluation. We employ ImageNette \citep{imagenette} as the dataset for this experiment, which is a subset of ImageNet consisting of $10$ categories. We learn $\tilde{{\mathbf{v}}}$ on the training dataset ($\approx 9500$ images) and perform evaluation on the validation set ($\approx 4000$ images).

To quantitatively evaluate learned $\tilde{\mathbf{v}}_\phi$, in \cref{fig:enc_ns} (a) we report the percentage of matching classifications with and without learned nullspace noise, and the mean squared error computed at the output probabilities (hereafter ``MSE confidence''). We consider a prediction to be matched if the assigned category for input is the same with and without adding the perturbation. By varying the regularization strength, we get noise vectors of different magnitude (\cref{fig:enc_ns} (b)), all being fairly benign to the model's predictions. However, if we randomly reset the vectors' direction by permuting their elements, the noise significant degrades the model's predictions. 

The experiment shows the feasibility of learning elements that approximately conform to our definition of generalized nullspace with good precision, and also indicates that at different magnitudes there are certain directions in the input space toward which the perturbation is fairly benign to the model. In Appendix E, we further empirically show that the learned noise vectors exhibit good properties under scalar multiplication and convex combinations within certain scope of parameters, similar to the closure property of a vector space.

\section{Nullspace Noise Augmented Finetuning}
\label{sec:ns_training}
In this section, we investigate the application of the synthesized nullspace noise. As we discussed previously, the model is weakly invariant to the learnt noise ($\mathfrak{I}$ in \cref{n_opt}) and the set as a result of this relaxed notion is an approximate nullspace. To more accurately quantify this, we define the \emph{$\epsilon$-Approximate Generalized Nullspace} as follows (later called ``$\epsilon$-approximate nullspace'' or ``approximate nullspace'' for brevity):
\begin{equation}
    \label{eq:eps_null}
    \tilde{\mathcal{N}}_\phi(\epsilon) = \{\tilde{\mathbf{v}} | \mathbb{E}_{\mathbf{u} \in \mathcal{D}} \,\lVert f(\mathbf{u}+\tilde{\mathbf{v}})-f(\mathbf{u}) \rVert \leq \epsilon \},
\end{equation}
where $f(\cdot) = \operatorname{Softmax}(f_\psi(f_\phi^{0}(\cdot)))$. It is easy to verify that $\forall \epsilon > 0, \mathbf{0} \in \tilde{\mathcal{N}}_\phi(\epsilon)$, and that $\forall \epsilon_2 > \epsilon_1 > 0$, $\tilde{\mathcal{N}}_\phi(\epsilon_1) \subseteq \tilde{\mathcal{N}}_\phi(\epsilon_2) $. 

We believe that the existence of approximate noise vectors is a property of the model. As these vectors
exhibit relaxed invariance, we also believe that they play a key role in model's inherent robustness under a variety of distribution shifts. Hence, if we can further improve invariance on approximate nullspace elements, we can potentially make the model more robust. With this belief, \textbf{we propose to fine-tune a pre-trained ViT with the learnt nullspace noise vector as an added (encoder level) input perturbation.} The motivation behind this is to enlarge the (approximate nullspace) set of noise vectors to which the model is invariant. 

Formally, we employ a bi-level optimization approach, where the inner problem finds the best noise vector and the outer problem finds the model that is the most tolerant to such noise, as shown below.
\begin{equation}
\begin{gathered} 
 \label{eq:n_opt}
    \min_{\phi} \quad \mathbb{E}_{\mathbf{u} \in \mathcal{D}}\, \ell(f_\psi(f_\phi^{0}(\mathbf{u} + \tilde{\mathbf{v}}_\phi^*)), \mathbf{y}) \\
    \text{where} \quad \tilde{\mathbf{v}}_{\phi}^* = \arg\max_{\tilde{\mathbf{v}}} \, \lVert \tilde{\mathbf{v}} \rVert 
    \ \quad \text{s.t.} \;\;\tilde{\mathbf{v}} \in \mathcal{N}_\phi(\epsilon).
\end{gathered}
\end{equation}
Here, $\ell(\cdot)$ is the cross-entropy loss. While this optimization problem can also be solved in different ways, we use an efficient heuristic: we initialize the noise with a large enough sampling limit, minimize $\mathcal{L_\phi}(\tilde{\mathbf{v}})$ by gradient descent according to the loss function in Equation~\ref{eq:ns_loss}, and early stop it as soon as it enters $\tilde{\mathcal{N}}_\phi(\epsilon)$, as shown in Equation~\ref{eq:ns_update}.
\begin{gather}
\mathcal{L_\phi}(\tilde{\mathbf{v}}) = \;\mathbb{E}_{\mathbf{u} \in \mathcal{D}} \,\lVert f_\psi(f_\phi^{0}(\mathbf{u}+\tilde{\mathbf{v}}))- f_\psi^{0}(f_\phi(\mathbf{u}))\rVert \, \label{eq:ns_loss} \\
\hat{\mathbf{v}}^* = \;\operatorname{SGD}(\mathcal{L_\phi}(\tilde{\mathbf{v}}), \tilde{\mathbf{v}}_0, \epsilon).
\label{eq:ns_update}
\end{gather}
Here, $\hat{\mathbf{v}}_\phi^*$ is the approximate solution for $\tilde{\mathbf{v}}_\phi^*$, $\operatorname{SGD}(\mathcal{L_\phi}(\tilde{\mathbf{v}}), \tilde{\mathbf{v}}_0, \epsilon)$ denotes the gradient descent algorithm that minimizes the loss $\mathcal{L_\phi}(\tilde{\mathbf{v}})$ starting from its initial value $\tilde{\mathbf{v}}_0$ until it satisfies the condition $\mathcal{L_\phi}(\tilde{\mathbf{v}}) < \epsilon$. 
The noise norm starts from a large value and gets gradually reduced during the process. When early stopping is triggered, we obtain noise vectors that are close to the boundary of the $\epsilon$-approximate nullspace. For more details of our method, please refer to \cref{alg_ns} in \cref{sec:app_alg}.

\section{Experiments}
\label{sec:experiments}
\subsection{Implementation Details}

In this section, we conduct evaluation of our nullspace augmented finetuning method (Section~\ref{sec:ns_training}) on several benchmarks. By making the model more tolerant to noise in the $\epsilon$-approximate nullspace, we hope to expand the nullspace itself and observe its effect on the model's robustness under different settings.

Starting from a pretrained model, we use the $\epsilon$-approximate nullspace noise as data augmentation to fine-tune the model. The noise is generated every 40 training steps according to \cref{eq:ns_update} with an $\epsilon$ of 0.03. The experiment was done within one epoch of training on the ImageNet-1k \citep{deng2009imagenet} dataset. We used the vanilla \texttt{ViT-small} and \texttt{ViT-base} models, and \texttt{ViT-base(DAT)} which is the current SOTA on ImageNet-C dataset on the EasyRobust benchmark\footnote{https://github.com/alibaba/easyrobust}, trained using Discrete Adversarial Training~\citep{mao2022enhance}. We evaluated the model performance in a wide range of settings to test its performance on the i.i.d dataset, under adversarial attacks and distribution shifts. For adversarial attacks we utilize FGSM \citep{adv_ex_0}, DamageNet \citep{damagenet}, PatchFool~\citep{fu2022patchfool} and CW~\citep{towards2017carlini}. Among them, FGSM and CW are gradient-based white-box attacks, DamageNet consists of pre-generated adversarial examples, and PatchFool targets localized, adversarial patches of an image.
For distribution shift we employ ImageNet-C \citep{hendrycks2018benchmarking}, ImageNet-A \citep{hendrycks2021natural}, ImageNet-V2 \citep{recht2019imagenet}, Imagenet-R \citep{Hendrycks_2021_ICCV}, ImageNet-Sketch \citep{wang2019learning} and Stylized-Imagenet \citep{geirhos2018imagenet}. ImageNet-C consists of validation images modified by applying corruptions in the form of weather effects, noises, etc. ImageNet-A applies the imagenet objects in hard contexts. ImageNet-R and ImageNet-Sketch consist of imagenet categories in different art forms. ImageNet-Stylized applies texture transfer onto the ImageNet validation images to create shape-texture contradictions. 

We use the EasyRobust library~\citep{mao2022easyrobust} for code implementation and the checkpoints of \texttt{ViT-base(DAT)}. For more implementation details please see our supplementary document.

\begin{table*}[!t]

\centering
\captionsetup{position=top}
\caption{\small \textbf{Effect of our nullspace augmented finetuning (NS) method on different models evaluated on multiple benchmark datasets.} Excluding DAT, vanilla ViT-S and ViT-B, the values for the baselines are directly reported from the corresponding papers. For DAT, we report the reproduced results following their evaluation setting.}
\setlength{\tabcolsep}{1.5mm}{
\resizebox{\textwidth}{!}{
\begin{tabular}{lcccccccccccc}
\toprule
 \multirow{2}{*}{Methods} & \multirow{2}{*}{Clean} & \multicolumn{4}{c}{Adversarial Robustness} & \multicolumn{6}{c}{Out of Distribution Robustness} & \multirow{2}{*}{Average}\\
 & & PatchFool & CW & FGSM & DamageNet & A & C$\downarrow$ & V2 & R & Sketch & Stylized & \\
 \cmidrule(lr){3-6} \cmidrule(lr){7-12}
\midrule
ViT-S & 74.19 & 0.68 & 4.63 & 13.79 & 29.82 & 16.35 & 71.13 & 62.51 & 34.67 & 14.26 & 12.15 & 26.54\\
ViT-S + NS (ours) & \textbf{77.47} & \textbf{19.10} & \textbf{9.37} & \textbf{25.95} & \textbf{32.43} & \textbf{20.77} & \textbf{55.98} & \textbf{66.5} & \textbf{41.61} & \textbf{25.67} & \textbf{16.02} & \textbf{34.45}\\
\midrule
ViT-B & 77.68 & 15.92 & 12.54 & 25.65 & 38.69 & 23.88 & 62.16 & 66.05 & 41.63 & 16.31 & 17.97 & 34.01\\
ViT-B + MixUp ~\citep{zhang2018mixup} & 77.80 & -- & -- & -- & -- & 12.20 & 61.80 & -- & 34.90 & -- & -- & --  \\
ViT-B + RandAugment~\citep{randaug} & 79.10 & -- & -- & -- & -- & -- & 43.60 & -- & 23.00 & -- & -- & --  \\
ViT-B + PR \citep{PatchRobust_22} & 78.20 & -- & -- & -- & -- & -- & 47.60 & -- & 21.40 & -- & -- & --  \\
ViT-B + RandAugment + PR & 79.30 & -- & -- & -- & -- & -- & 43.60 & -- & 23.80 & -- & -- & --  \\
ViT-B + AugMix~\citep{hendrycks_2020augmix} & 78.80 & -- & -- & -- & -- & -- & 42.20 & -- & 24.90 & -- & -- & --  \\
ViT-B + AugMix + PR & 79.30 & -- & -- & -- & -- & -- & \textbf{41.60} & -- & 25.70 & -- & -- & --  \\
ViT-B + SAM~\citep{chen2022when} & 79.90 & -- & -- & -- & -- & -- & 43.50 & 67.50 & 26.40 & -- & -- & --  \\
RobustViT-B~\citep{robustvit} & 80.40 & -- & -- & -- & -- & 23.00 & -- & 69.80 & 35.40 & \textbf{35.80} & -- & -- \\
\rowcolor[HTML]{E3F2FD}[8pt]
ViT-B + NS & \textbf{81.42} & \textbf{23.52} & 
\textbf{14.23} & \textbf{36.50} & \textbf{40.44} & \textbf{24.55} & 47.82 & \textbf{70.25} & \textbf{44.85} & 26.35 & \textbf{19.02} & \textbf{39.39}\\
\midrule
 ViT-B + DAT\citep{mao2022enhance} & \textbf{81.47} &  22.64 & 23.59 & 48.80 & 43.31 & 23.83 & 45.95 & \textbf{70.24} & \textbf{48.68} & 36.94 & \textbf{23.99} & 43.41\\
\rowcolor[HTML]{E3F2FD}[8pt]
 ViT-B + DAT + NS & 81.33 & \textbf{24.14} & \textbf{23.61} & \textbf{48.98} & \textbf{43.67} & \textbf{24.22} & \textbf{45.91} & 70.14 & 48.48 & \textbf{37.25} & 23.87 & \textbf{43.61}\\
\bottomrule
\end{tabular}
}
}
\label{tab:baseline}
\end{table*}
\subsection{Experiment: Robustness Evaluation}

We evaluated the effect of nullspace finetuning to improve the robustness of vision transformers under different settings. We used the official mCE score as the evaluation metric for ImageNet-C, where a lower mCE indicates better robustness, and we used the accuracy score for all other settings. We used $100-\textrm{mCE}$ before taking the average in all settings. 

The result in Table~\ref{tab:baseline} shows that our nullspace finetuning method consistently improves the robustness of models under distribution shifts and adversarial attacks, yielding a large gain in average performance for the vanilla \texttt{ViT-small} and \texttt{ViT-base model}, and slightly outperforms various baselines consistently while also slightly outperforming DAT. 
This not only shows that our nullspace finetuning method is effective but also validates our previous hypothesis about the connection between the tolerance to nullspace and the robustness of transformer models.
\subsection{Experiment: Adversarial Finetuning}
In this experiment, we compare our method with fine-tuning using two PGD adversarial training methods, Madry~\citep{madry} and TRADES~\citep{trades} on the ViT-S model. TRADES, in each training iteration, generates adversarial examples using PGD and updates the model's parameters to minimize the worst-case loss on these adversarial examples while also minimizing the standard classification loss on clean data. Madry, on the other hand, focuses exclusively on minimizing the worst-case loss on adversarial examples.
\begin{table}[t]
    \centering
         \caption{\small \textbf{Comparison of our NS method with PGD-based adversarial robustness methods of Madry and TRADES.} We report the performance for a ViT-S model.}
    \small
\setlength{\tabcolsep}{2.75mm}  
\begin{tabular}{cccccccccc}
    \toprule
        Method  & clean  & FGSM  & DamageNet  & A  & C ($\downarrow$)  & V2  & R  & Sketch  & Stylized \\ 
        \midrule
        ViT-S & 74.19 & 13.79 & 29.82 & 16.35 & 71.13 & 62.51 & 34.67 & 14.26 & 12.15 \\
        \midrule
        Madry  & 70.53  & \textbf{39.37}  & \textbf{49.91}  & 9.37  & 81.74  & 58.88  & 39.04  & 21.36  & 10.76 \\ 
        TRADES  & 74.02  & 38.85  & 36.28  & 16.53  & 73.11  & 63.37  & 40.86  & \textbf{26.43}  & 13.22 \\ 
        \rowcolor[HTML]{E3F2FD}[10pt]
        NS  & \textbf{77.47}  & 25.95  & 32.43  & \textbf{20.77}  & \textbf{55.98}  & \textbf{66.5}  & \textbf{41.61}  & 25.67  & \textbf{16.02} \\ 
    \bottomrule
    \end{tabular}
  \label{tab:exp_pgd}
\end{table}
In \cref{tab:exp_pgd}, we observe that Madry and TRADES provide better performance for adversarial evaluation. This is expected as the methods are catered for improving adversarial robustness. However, this exclusivity leads to relatively poorer performance in a wider benchmark evaluation. Compared to our method, Madry and TRADES perform considerably lower in the natural OOD setting. 

\subsection{Enlarged Approximate Nullspace}
\begin{wrapfigure}{r}{0.48\textwidth}
  \vspace{-4.7mm}
  \centering
  \includegraphics[width=1\linewidth]{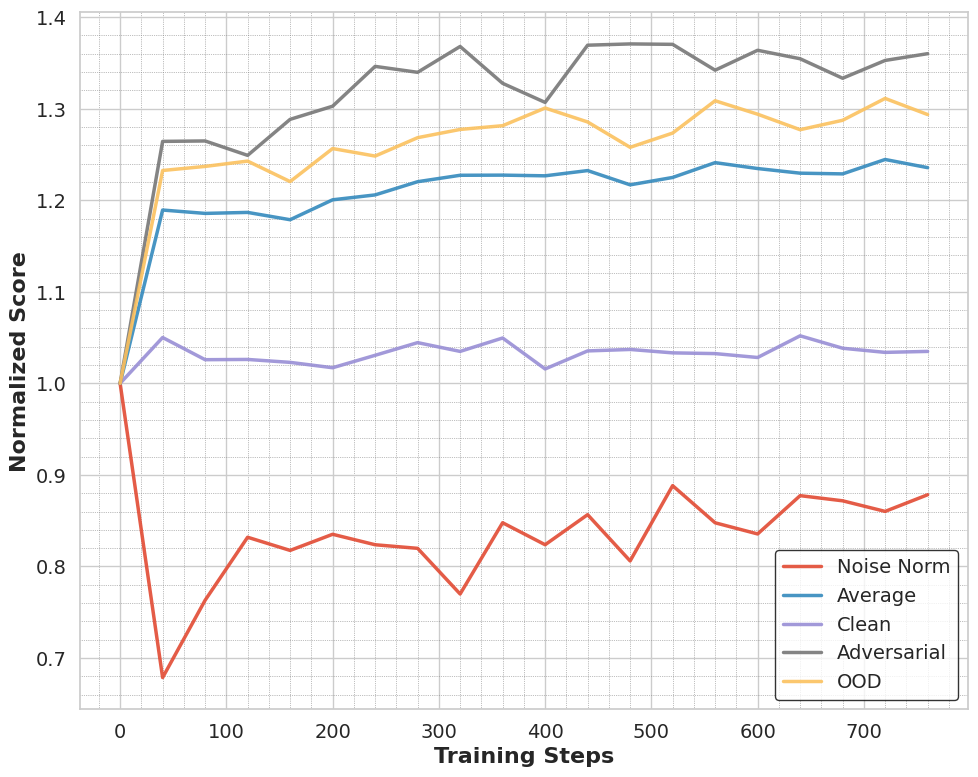}
  \caption{\small \textbf{Change trend of multiple metrics with training steps.} "Adversarial" is the average performance of the 4 adv. robustness settings, "OOD" is the average score on the six OOD datasets, and "avg" is the total average. All values are divided by their initial values to show the trend more clearly. }
  \label{fig:trend}
\end{wrapfigure}
To gain more insight about the dynamics of our nullspace finetuning method, we monitor the $l_2$ norm of the learned noise and various performance metrics during the training, 
as shown in Fig.~\ref{fig:trend}. 
Before the nullspace finetuning, it was hard to optimize the noise into the $\epsilon$ region even with increased training, so the norm started with a high value. 
As the training starts, we find that the noise was always able to enter the $\epsilon$ region. In Appendix~\ref{sec:validate_mse}, we show the MSE probability of the learned noise vectors after each round of noise learning, which were all smaller than $\epsilon$. More importantly, the norm of the learned noise gradually increases along the process of model fine-tuning. 
The fluctuation may have mainly resulted from the randomness in mini-batches and the optimization dynamics. 
The model allows for noises with larger and larger norms to be within $\epsilon$-approximate, which informally suggests an enlarging $\epsilon$-approximate nullspace. 
Accompanied by the trend is the increase in robustness scores in both OOD and adversarial settings, which corroborates our findings. 

\begin{table}[t!]
\centering
\small
\caption{\small \textbf{Impact of $\epsilon$ on the final performance}. Moreover, we also compare our NS method against random $\epsilon$ noise based finetuning.}
\setlength{\tabcolsep}{2.75mm}
    \begin{tabular}{cccccccccc}
    \toprule
        $\epsilon$ & Finetuning & FGSM  & DamageNet  & A  & C ($\downarrow$)  & V2  & R  & Sketch  & Stylized \\ 
        \midrule
        \multirow{2}{*}{0.01}  & NS  & \textbf{26.04}  & \textbf{33.65}  & \textbf{20.45}  & 56.26  & \textbf{66.47}  & \textbf{41.4}  & \textbf{23.34}  & \textbf{15.85} \\ 
          & Random  & 21.54  & 28.81  & 17.07  & \textbf{55.13}  & 61.98  & 34.97  & 14.43  & 12.14 \\ \midrule
        \multirow{2}{*}{0.03}  & NS  & \textbf{25.95}  & \textbf{32.43}  & \textbf{20.77}  & 55.98  & \textbf{66.5}  & \textbf{41.61}  & \textbf{25.67}  & \textbf{16.02} \\
          & Random  & 23.18  & 29.61  & 16.91  & \textbf{54.68}  & 62.2  & 35.05  & 14.77  & 12.34 \\ \midrule
        \multirow{2}{*}{0.1}  & NS  & \textbf{25.38}  & \textbf{33.09}  & \textbf{20.16}  & 56.41  & \textbf{66.47}  & \textbf{40.42}  & \textbf{22.66}  & \textbf{15.78} \\ 
        & Random  & 23.93  & 30.56  & 16.47  & \textbf{54.52}  & 62.48  & 34.66  & 14.99  & 12.35 \\ \bottomrule
    \end{tabular}
    \label{tab:ablation}
\end{table}

\subsection{Ablation Study}
We conduct an extensive study to analyse the performance of our method under choice of $\epsilon$. Furthermore, we also compare our approach with a simple baseline of using an $\epsilon$ noise sampled from a Gaussian distribution.

From Table \ref{tab:ablation}, we can infer that the nullspace noise based finetuning is relatively robust to the choice of $\epsilon$. Moreover, compared to using randomly generated $\epsilon$-noise, our nullspace based training provides significant performance boost. This observation stands across different values of $\epsilon$.

\section{Discussion}
\label{sec:discussion}
\textbf{Applications in Model Patenting}
In addition to the applications we discussed, 
we consider another potential usage of our findings is to 
patent a ViT after a model is trained, 
as the nullspace will be unique property of any set of weights 
of certain ViT architectures. 
Different from the existing line of research in model watermarking \citep{adi2018turning,darvish2019deepsigns,le2020adversarial}, 
the patenting through nullspace will not require any additional steps during training, 
although will face limited usage scenarios in comparison. 

\textbf{Applications in Image Watermarking}
Using the nullspace noise, it is possible to apply signatures onto input images without harming the output or operability of the networks. In the supplementary document, we present the cases where certain marks in form of nullspace noise can be superimposed on any desired input image. 

\textbf{Potential Limitation about the Nullspace Approximation}
Different from the nullspace defined in linear algebra, 
the nullspace of the entire ViT can only be approximated because of 
the non-linearity in the network architecture. 
However, it is worthy mentioning that
we can still calculate the exact nullspace of ViT if we only consider the patch embedding layer, 
through which, our results will qualitatively deliver the same message, 
but with quantitative differences. 

\section{Conclusion}
\label{sec:conclusion}
In this work, we have explored the concept of nullspace in Vision Transformers (ViTs) to understand their robustness. Our findings demonstrate that a non-trivial nullspace indeed exists for Vision Transformers, a direct consequence of the patch embedding layer. This discovery implies that there are elements that, when added to an input, do not affect the output of the network, potentially offering an explanation for the robustness exhibited by ViTs. Moreover, we have extended the definition of nullspace, preserving a property that implies invariance of a mapping's output to input perturbations, and empirically identified a space that exhibits such property within the input space of the non-linear transformer encoder. By linking the presence of nullspace with our extended definition to the general robustness of a network, we were able to devise a new approach to improve the robustness of ViTs. Our empirical results suggest that fine-tuning ViTs with the learnt nullspace noise can significantly enhance their robustness to a variety of robustness benchmarks. 

This study offers a new perspective to the robustness of vision transformers. We believe these findings can assist in furthering the robustness of ViTs, potentially facilitating advancements in the development of more resilient models. Looking forward, there is more to explore in this direction. Future research could focus on the development of efficient algorithms for learning nullspace and investigate its presence in other architectures and layers of deep neural networks.

\newpage
\bibliographystyle{abbrvnat}
\bibliography{main}

\begin{thebibliography}{78}
\providecommand{\natexlab}[1]{#1}
\providecommand{\url}[1]{\texttt{#1}}
\expandafter\ifx\csname urlstyle\endcsname\relax
  \providecommand{\doi}[1]{doi: #1}\else
  \providecommand{\doi}{doi: \begingroup \urlstyle{rm}\Url}\fi

\bibitem[Abdelpakey and Shehata(2020)]{nullnet}
M.~H. Abdelpakey and M.~S. Shehata.
\newblock Nullspacenet: Nullspace convoluional neural network with differentiable loss function.
\newblock \emph{CoRR}, abs/2004.12058, 2020.
\newblock URL \url{https://arxiv.org/abs/2004.12058}.

\bibitem[Abnar and Zuidema(2020)]{rollout}
S.~Abnar and W.~Zuidema.
\newblock Quantifying attention flow in transformers.
\newblock In \emph{Proceedings of the 58th Annual Meeting of the Association for Computational Linguistics}, pages 4190--4197, Online, July 2020. Association for Computational Linguistics.
\newblock \doi{10.18653/v1/2020.acl-main.385}.
\newblock URL \url{https://aclanthology.org/2020.acl-main.385}.

\bibitem[Adi et~al.(2018)Adi, Baum, Cisse, Pinkas, and Keshet]{adi2018turning}
Y.~Adi, C.~Baum, M.~Cisse, B.~Pinkas, and J.~Keshet.
\newblock Turning your weakness into a strength: Watermarking deep neural networks by backdooring.
\newblock In \emph{27th USENIX Security Symposium (USENIX Security 18)}, pages 1615--1631, 2018.

\bibitem[Al-Haj(2007)]{al2007combined}
A.~Al-Haj.
\newblock Combined dwt-dct digital image watermarking.
\newblock \emph{Journal of computer science}, 3\penalty0 (9):\penalty0 740--746, 2007.

\bibitem[Ali et~al.(2021)Ali, Touvron, Caron, Bojanowski, Douze, Joulin, Laptev, Neverova, Synnaeve, Verbeek, et~al.]{ali2021xcit}
A.~Ali, H.~Touvron, M.~Caron, P.~Bojanowski, M.~Douze, A.~Joulin, I.~Laptev, N.~Neverova, G.~Synnaeve, J.~Verbeek, et~al.
\newblock Xcit: Cross-covariance image transformers.
\newblock \emph{Advances in neural information processing systems}, 34:\penalty0 20014--20027, 2021.

\bibitem[Bai et~al.(2021)Bai, Mei, Yuille, and Xie]{NEURIPS2021_bai}
Y.~Bai, J.~Mei, A.~L. Yuille, and C.~Xie.
\newblock Are transformers more robust than cnns?
\newblock In M.~Ranzato, A.~Beygelzimer, Y.~Dauphin, P.~Liang, and J.~W. Vaughan, editors, \emph{Advances in Neural Information Processing Systems}, volume~34, pages 26831--26843. Curran Associates, Inc., 2021.
\newblock URL \url{https://proceedings.neurips.cc/paper/2021/file/e19347e1c3ca0c0b97de5fb3b690855a-Paper.pdf}.

\bibitem[Berghel(1998)]{berghel1998digital}
H.~Berghel.
\newblock Digital watermarking makes its mark.
\newblock \emph{Networker}, 2\penalty0 (4):\penalty0 30--39, 1998.

\bibitem[Bhat et~al.(2010)Bhat, Sengupta, and Das]{bhat2010adaptive}
V.~Bhat, I.~Sengupta, and A.~Das.
\newblock An adaptive audio watermarking based on the singular value decomposition in the wavelet domain.
\newblock \emph{Digital Signal Processing}, 20\penalty0 (6):\penalty0 1547--1558, 2010.

\bibitem[Carlini and Wagner(2017)]{towards2017carlini}
N.~Carlini and D.~Wagner.
\newblock Towards evaluating the robustness of neural networks.
\newblock In \emph{2017 IEEE Symposium on Security and Privacy (SP)}, pages 39--57, Los Alamitos, CA, USA, may 2017. IEEE Computer Society.
\newblock \doi{10.1109/SP.2017.49}.
\newblock URL \url{https://doi.ieeecomputersociety.org/10.1109/SP.2017.49}.

\bibitem[Chatzipantazis et~al.(2023)Chatzipantazis, Pertigkiozoglou, Daniilidis, and Dobriban]{chatzipantazis2023learning}
E.~Chatzipantazis, S.~Pertigkiozoglou, K.~Daniilidis, and E.~Dobriban.
\newblock Learning augmentation distributions using transformed risk minimization.
\newblock \emph{Transactions on Machine Learning Research}, 2023.
\newblock ISSN 2835-8856.
\newblock URL \url{https://openreview.net/forum?id=LRYtNj8Xw0}.

\bibitem[Chefer et~al.(2021)Chefer, Gur, and Wolf]{Chefer_2021_CVPR}
H.~Chefer, S.~Gur, and L.~Wolf.
\newblock Transformer interpretability beyond attention visualization.
\newblock In \emph{Proceedings of the IEEE/CVF Conference on Computer Vision and Pattern Recognition (CVPR)}, pages 782--791, June 2021.

\bibitem[Chefer et~al.(2022{\natexlab{a}})Chefer, Schwartz, and Wolf]{chefer2022optimizing}
H.~Chefer, I.~Schwartz, and L.~Wolf.
\newblock Optimizing relevance maps of vision transformers improves robustness.
\newblock In A.~H. Oh, A.~Agarwal, D.~Belgrave, and K.~Cho, editors, \emph{Advances in Neural Information Processing Systems}, 2022{\natexlab{a}}.
\newblock URL \url{https://openreview.net/forum?id=upuYKQiyxa_}.

\bibitem[Chefer et~al.(2022{\natexlab{b}})Chefer, Schwartz, and Wolf]{robustvit}
H.~Chefer, I.~Schwartz, and L.~Wolf.
\newblock Optimizing relevance maps of vision transformers improves robustness.
\newblock In S.~Koyejo, S.~Mohamed, A.~Agarwal, D.~Belgrave, K.~Cho, and A.~Oh, editors, \emph{NeurIPS}, volume~35, pages 33618--33632. Curran Associates, Inc., 2022{\natexlab{b}}.
\newblock URL \url{https://proceedings.neurips.cc/paper_files/paper/2022/file/d9fa720cf96f7c18ac4d9e04270f0bbf-Paper-Conference.pdf}.

\bibitem[Chen et~al.(2020)Chen, Dobriban, and Lee]{chen2020group}
S.~Chen, E.~Dobriban, and J.~H. Lee.
\newblock A group-theoretic framework for data augmentation.
\newblock \emph{The Journal of Machine Learning Research}, 21\penalty0 (1):\penalty0 9885--9955, 2020.

\bibitem[Chen et~al.(2022{\natexlab{a}})Chen, He, Sun, Yang, and Huang]{damagenet}
S.~Chen, Z.~He, C.~Sun, J.~Yang, and X.~Huang.
\newblock Universal adversarial attack on attention and the resulting dataset damagenet.
\newblock \emph{IEEE Transactions on Pattern Analysis and Machine Intelligence}, 44\penalty0 (4):\penalty0 2188--2197, 2022{\natexlab{a}}.
\newblock \doi{10.1109/TPAMI.2020.3033291}.

\bibitem[Chen et~al.(2022{\natexlab{b}})Chen, Hsieh, and Gong]{chen2022when}
X.~Chen, C.-J. Hsieh, and B.~Gong.
\newblock When vision transformers outperform resnets without pre-training or strong data augmentations.
\newblock In \emph{ICLR}, 2022{\natexlab{b}}.
\newblock URL \url{https://openreview.net/forum?id=LtKcMgGOeLt}.

\bibitem[Clarysse et~al.(2023)Clarysse, H{\"o}rrmann, and Yang]{clarysse2023why}
J.~Clarysse, J.~H{\"o}rrmann, and F.~Yang.
\newblock Why adversarial training can hurt robust accuracy.
\newblock In \emph{ICLR}, 2023.
\newblock URL \url{https://openreview.net/forum?id=-CA8yFkPc7O}.

\bibitem[Cubuk et~al.(2020)Cubuk, Zoph, Shlens, and Le]{randaug}
E.~D. Cubuk, B.~Zoph, J.~Shlens, and Q.~Le.
\newblock Randaugment: Practical automated data augmentation with a reduced search space.
\newblock In H.~Larochelle, M.~Ranzato, R.~Hadsell, M.~Balcan, and H.~Lin, editors, \emph{NeurIPS}, volume~33, pages 18613--18624. Curran Associates, Inc., 2020.
\newblock URL \url{https://proceedings.neurips.cc/paper_files/paper/2020/file/d85b63ef0ccb114d0a3bb7b7d808028f-Paper.pdf}.

\bibitem[Darvish~Rouhani et~al.(2019)Darvish~Rouhani, Chen, and Koushanfar]{darvish2019deepsigns}
B.~Darvish~Rouhani, H.~Chen, and F.~Koushanfar.
\newblock Deepsigns: An end-to-end watermarking framework for ownership protection of deep neural networks.
\newblock In \emph{Proceedings of the Twenty-Fourth International Conference on Architectural Support for Programming Languages and Operating Systems}, pages 485--497, 2019.

\bibitem[Deng et~al.(2009)Deng, Dong, Socher, Li, Li, and Fei-Fei]{deng2009imagenet}
J.~Deng, W.~Dong, R.~Socher, L.-J. Li, K.~Li, and L.~Fei-Fei.
\newblock Imagenet: A large-scale hierarchical image database.
\newblock In \emph{2009 IEEE conference on computer vision and pattern recognition}, pages 248--255. Ieee, 2009.

\bibitem[Dombrowski et~al.(2019)Dombrowski, Alber, Anders, Ackermann, M{\"u}ller, and Kessel]{dombrowski2019explanations}
A.-K. Dombrowski, M.~Alber, C.~Anders, M.~Ackermann, K.-R. M{\"u}ller, and P.~Kessel.
\newblock Explanations can be manipulated and geometry is to blame.
\newblock \emph{Advances in Neural Information Processing Systems}, 32, 2019.

\bibitem[Dosovitskiy et~al.(2021)Dosovitskiy, Beyer, Kolesnikov, Weissenborn, Zhai, Unterthiner, Dehghani, Minderer, Heigold, Gelly, Uszkoreit, and Houlsby]{dosovitskiy2020vit}
A.~Dosovitskiy, L.~Beyer, A.~Kolesnikov, D.~Weissenborn, X.~Zhai, T.~Unterthiner, M.~Dehghani, M.~Minderer, G.~Heigold, S.~Gelly, J.~Uszkoreit, and N.~Houlsby.
\newblock An image is worth 16x16 words: Transformers for image recognition at scale.
\newblock \emph{ICLR}, 2021.

\bibitem[Esser et~al.(2021)Esser, Rombach, and Ommer]{esser2021taming}
P.~Esser, R.~Rombach, and B.~Ommer.
\newblock Taming transformers for high-resolution image synthesis.
\newblock In \emph{Proceedings of the IEEE/CVF conference on computer vision and pattern recognition}, pages 12873--12883, 2021.

\bibitem[Fu et~al.(2022)Fu, Zhang, Wu, Wan, and Lin]{fu2022patchfool}
Y.~Fu, S.~Zhang, S.~Wu, C.~Wan, and Y.~Lin.
\newblock Patch-fool: Are vision transformers always robust against adversarial perturbations?
\newblock In \emph{International Conference on Learning Representations}, 2022.
\newblock URL \url{https://openreview.net/forum?id=28ib9tf6zhr}.

\bibitem[Geirhos et~al.(2018)Geirhos, Rubisch, Michaelis, Bethge, Wichmann, and Brendel]{geirhos2018imagenet}
R.~Geirhos, P.~Rubisch, C.~Michaelis, M.~Bethge, F.~A. Wichmann, and W.~Brendel.
\newblock Imagenet-trained cnns are biased towards texture; increasing shape bias improves accuracy and robustness.
\newblock In \emph{International Conference on Learning Representations}, 2018.

\bibitem[Ghorbani et~al.(2019)Ghorbani, Abid, and Zou]{ghorbani2019interpretation}
A.~Ghorbani, A.~Abid, and J.~Zou.
\newblock Interpretation of neural networks is fragile.
\newblock In \emph{Proceedings of the AAAI conference on artificial intelligence}, volume~33, pages 3681--3688, 2019.

\bibitem[Goggin et~al.(1992)Goggin, Gustafson, and Johnson]{mlp_null}
S.~D.~D. Goggin, K.~E. Gustafson, and K.~M. Johnson.
\newblock {Accessing the null space with nonlinear multilayer neural networks}.
\newblock In D.~W. Ruck, editor, \emph{Science of Artificial Neural Networks}, volume 1710, pages 308 -- 316. International Society for Optics and Photonics, SPIE, 1992.
\newblock \doi{10.1117/12.140097}.
\newblock URL \url{https://doi.org/10.1117/12.140097}.

\bibitem[Goodfellow et~al.(2015)Goodfellow, Shlens, and Szegedy]{adv_ex_0}
I.~J. Goodfellow, J.~Shlens, and C.~Szegedy.
\newblock Explaining and harnessing adversarial examples.
\newblock In Y.~Bengio and Y.~LeCun, editors, \emph{3rd International Conference on Learning Representations, {ICLR} 2015, San Diego, CA, USA, May 7-9, 2015, Conference Track Proceedings}, 2015.
\newblock URL \url{http://arxiv.org/abs/1412.6572}.

\bibitem[Gowal et~al.(2021)Gowal, Qin, Uesato, Mann, and Kohli]{gowal2021uncovering}
S.~Gowal, C.~Qin, J.~Uesato, T.~Mann, and P.~Kohli.
\newblock Uncovering the limits of adversarial training against norm-bounded adversarial examples, 2021.

\bibitem[Hendrycks and Dietterich(2018)]{hendrycks2018benchmarking}
D.~Hendrycks and T.~Dietterich.
\newblock Benchmarking neural network robustness to common corruptions and perturbations.
\newblock In \emph{International Conference on Learning Representations}, 2018.

\bibitem[Hendrycks et~al.(2020)Hendrycks, Mu, Cubuk, Zoph, Gilmer, and Lakshminarayanan]{hendrycks_2020augmix}
D.~Hendrycks, N.~Mu, E.~D. Cubuk, B.~Zoph, J.~Gilmer, and B.~Lakshminarayanan.
\newblock Augmix: A simple method to improve robustness and uncertainty under data shift.
\newblock In \emph{ICLR}, 2020.
\newblock URL \url{https://openreview.net/forum?id=S1gmrxHFvB}.

\bibitem[Hendrycks et~al.(2021{\natexlab{a}})Hendrycks, Basart, Mu, Kadavath, Wang, Dorundo, Desai, Zhu, Parajuli, Guo, Song, Steinhardt, and Gilmer]{Hendrycks_2021_ICCV}
D.~Hendrycks, S.~Basart, N.~Mu, S.~Kadavath, F.~Wang, E.~Dorundo, R.~Desai, T.~Zhu, S.~Parajuli, M.~Guo, D.~Song, J.~Steinhardt, and J.~Gilmer.
\newblock The many faces of robustness: A critical analysis of out-of-distribution generalization.
\newblock In \emph{Proceedings of the IEEE/CVF International Conference on Computer Vision (ICCV)}, pages 8340--8349, October 2021{\natexlab{a}}.

\bibitem[Hendrycks et~al.(2021{\natexlab{b}})Hendrycks, Zhao, Basart, Steinhardt, and Song]{hendrycks2021natural}
D.~Hendrycks, K.~Zhao, S.~Basart, J.~Steinhardt, and D.~Song.
\newblock Natural adversarial examples.
\newblock In \emph{Proceedings of the IEEE/CVF Conference on Computer Vision and Pattern Recognition}, pages 15262--15271, 2021{\natexlab{b}}.

\bibitem[Heo et~al.(2019)Heo, Joo, and Moon]{heo2019fooling}
J.~Heo, S.~Joo, and T.~Moon.
\newblock Fooling neural network interpretations via adversarial model manipulation.
\newblock \emph{Advances in Neural Information Processing Systems}, 32, 2019.

\bibitem[Hounie et~al.(2023)Hounie, Chamon, and Ribeiro]{inv_learn}
I.~Hounie, L.~F.~O. Chamon, and A.~Ribeiro.
\newblock Automatic data augmentation via invariance-constrained learning.
\newblock In A.~Krause, E.~Brunskill, K.~Cho, B.~Engelhardt, S.~Sabato, and J.~Scarlett, editors, \emph{ICML}, volume 202 of \emph{Proceedings of Machine Learning Research}, pages 13410--13433. PMLR, 23--29 Jul 2023.
\newblock URL \url{https://proceedings.mlr.press/v202/hounie23a.html}.

\bibitem[Howard(2018)]{imagenette}
J.~Howard.
\newblock Imagenette, 2018.
\newblock URL \url{https://github.com/fastai/imagenette/}.

\bibitem[Kirillov et~al.(2023)Kirillov, Mintun, Ravi, Mao, Rolland, Gustafson, Xiao, Whitehead, Berg, Lo, et~al.]{kirillov2023segment}
A.~Kirillov, E.~Mintun, N.~Ravi, H.~Mao, C.~Rolland, L.~Gustafson, T.~Xiao, S.~Whitehead, A.~C. Berg, W.-Y. Lo, et~al.
\newblock Segment anything.
\newblock \emph{arXiv preprint arXiv:2304.02643}, 2023.

\bibitem[Kwak and Hong(2004)]{kwak04}
J.~H. Kwak and S.~Hong.
\newblock \emph{Linear Algebra}.
\newblock Birkh\"{a}user, Boston, MA, 2004.
\newblock \doi{10.1007/978-0-8176-8194-4}.

\bibitem[Le~Merrer et~al.(2020)Le~Merrer, Perez, and Tr{\'e}dan]{le2020adversarial}
E.~Le~Merrer, P.~Perez, and G.~Tr{\'e}dan.
\newblock Adversarial frontier stitching for remote neural network watermarking.
\newblock \emph{Neural Computing and Applications}, 32\penalty0 (13):\penalty0 9233--9244, 2020.

\bibitem[Li et~al.(2020)Li, Hu, Wang, Hospedales, Robertson, and Yang]{li2020dada}
Y.~Li, G.~Hu, Y.~Wang, T.~M. Hospedales, N.~M. Robertson, and Y.~Yang.
\newblock {DADA:} differentiable automatic data augmentation.
\newblock 2020.

\bibitem[Li et~al.(2022)Li, Yao, Pan, and Mei]{li2022contextual}
Y.~Li, T.~Yao, Y.~Pan, and T.~Mei.
\newblock Contextual transformer networks for visual recognition.
\newblock \emph{IEEE Transactions on Pattern Analysis and Machine Intelligence}, 2022.

\bibitem[Liu et~al.(2023{\natexlab{a}})Liu, Chaudhary, and Wang]{liu2023towards}
H.~Liu, M.~Chaudhary, and H.~Wang.
\newblock Towards trustworthy and aligned machine learning: A data-centric survey with causality perspectives.
\newblock \emph{arXiv preprint arXiv:2307.16851}, 2023{\natexlab{a}}.

\bibitem[Liu et~al.(2023{\natexlab{b}})Liu, Chaudhary, and Wang]{liu2023trustworthy}
H.~Liu, M.~Chaudhary, and H.~Wang.
\newblock Towards trustworthy and aligned machine learning: A data-centric survey with causality perspectives, 2023{\natexlab{b}}.

\bibitem[Liu et~al.(2022)Liu, Liu, Zhou, Li, and Liu]{liu2022tokenmix}
J.~Liu, B.~Liu, H.~Zhou, H.~Li, and Y.~Liu.
\newblock Tokenmix: Rethinking image mixing for data augmentation in vision transformers.
\newblock In \emph{European Conference on Computer Vision}, pages 455--471. Springer, 2022.

\bibitem[Liu et~al.(2021)Liu, Lin, Cao, Hu, Wei, Zhang, Lin, and Guo]{liu2021Swin}
Z.~Liu, Y.~Lin, Y.~Cao, H.~Hu, Y.~Wei, Z.~Zhang, S.~Lin, and B.~Guo.
\newblock Swin transformer: Hierarchical vision transformer using shifted windows.
\newblock In \emph{Proceedings of the IEEE/CVF International Conference on Computer Vision (ICCV)}, 2021.

\bibitem[Loshchilov and Hutter(2019)]{loshchilov2018decoupled}
I.~Loshchilov and F.~Hutter.
\newblock Decoupled weight decay regularization.
\newblock In \emph{International Conference on Learning Representations}, 2019.
\newblock URL \url{https://openreview.net/forum?id=Bkg6RiCqY7}.

\bibitem[Lyle et~al.(2020)Lyle, van~der Wilk, Kwiatkowska, Gal, and Bloem-Reddy]{lyle2020benefits}
C.~Lyle, M.~van~der Wilk, M.~Kwiatkowska, Y.~Gal, and B.~Bloem-Reddy.
\newblock On the benefits of invariance in neural networks.
\newblock \emph{arXiv preprint arXiv:2005.00178}, 2020.

\bibitem[Madry et~al.(2018{\natexlab{a}})Madry, Makelov, Schmidt, Tsipras, and Vladu]{madry}
A.~Madry, A.~Makelov, L.~Schmidt, D.~Tsipras, and A.~Vladu.
\newblock Towards deep learning models resistant to adversarial attacks.
\newblock In \emph{ICLR}. OpenReview.net, 2018{\natexlab{a}}.
\newblock URL \url{https://openreview.net/forum?id=rJzIBfZAb}.

\bibitem[Madry et~al.(2018{\natexlab{b}})Madry, Makelov, Schmidt, Tsipras, and Vladu]{madry2018towards}
A.~Madry, A.~Makelov, L.~Schmidt, D.~Tsipras, and A.~Vladu.
\newblock Towards deep learning models resistant to adversarial attacks.
\newblock In \emph{International Conference on Learning Representations}, 2018{\natexlab{b}}.
\newblock URL \url{https://openreview.net/forum?id=rJzIBfZAb}.

\bibitem[Mahmood et~al.(2021)Mahmood, Mahmood, and Van~Dijk]{mahmood2021robustness}
K.~Mahmood, R.~Mahmood, and M.~Van~Dijk.
\newblock On the robustness of vision transformers to adversarial examples.
\newblock In \emph{Proceedings of the IEEE/CVF International Conference on Computer Vision}, pages 7838--7847, 2021.

\bibitem[Mao et~al.(2022{\natexlab{a}})Mao, Chen, Duan, Zhu, Qi, Ye, Li, Zhang, and Xue']{mao2022enhance}
X.~Mao, Y.~Chen, R.~Duan, Y.~Zhu, G.~Qi, S.~Ye, X.~Li, R.~Zhang, and H.~Xue'.
\newblock Enhance the visual representation via discrete adversarial training.
\newblock In A.~H. Oh, A.~Agarwal, D.~Belgrave, and K.~Cho, editors, \emph{Advances in Neural Information Processing Systems}, 2022{\natexlab{a}}.
\newblock URL \url{https://openreview.net/forum?id=qtZac7A3-F}.

\bibitem[Mao et~al.(2022{\natexlab{b}})Mao, Chen, Li, Qi, Duan, Zhang, and Xue]{mao2022easyrobust}
X.~Mao, Y.~Chen, X.~Li, G.~Qi, R.~Duan, R.~Zhang, and H.~Xue.
\newblock Easyrobust: A comprehensive and easy-to-use toolkit for robust computer vision.
\newblock \url{https://github.com/alibaba/easyrobust}, 2022{\natexlab{b}}.

\bibitem[Moosavi-Dezfooli et~al.(2016)Moosavi-Dezfooli, Fawzi, and Frossard]{adv_ex_1}
S.-M. Moosavi-Dezfooli, A.~Fawzi, and P.~Frossard.
\newblock Deepfool: A simple and accurate method to fool deep neural networks.
\newblock In \emph{2016 IEEE Conference on Computer Vision and Pattern Recognition (CVPR)}, pages 2574--2582, 2016.
\newblock \doi{10.1109/CVPR.2016.282}.

\bibitem[M\"uller and Hutter(2021)]{Muller_2021_ICCV}
S.~G. M\"uller and F.~Hutter.
\newblock Trivialaugment: Tuning-free yet state-of-the-art data augmentation.
\newblock In \emph{ICCV}, pages 774--782, October 2021.

\bibitem[Nguyen et~al.(2015)Nguyen, Yosinski, and Clune]{adv_ex_2}
A.~Nguyen, J.~Yosinski, and J.~Clune.
\newblock Deep neural networks are easily fooled: High confidence predictions for unrecognizable images.
\newblock In \emph{2015 IEEE Conference on Computer Vision and Pattern Recognition (CVPR)}, pages 427--436, 2015.
\newblock \doi{10.1109/CVPR.2015.7298640}.

\bibitem[Paul and Chen(2022)]{paul2021vision}
S.~Paul and P.-Y. Chen.
\newblock Vision transformers are robust learners.
\newblock \emph{Proceedings of the AAAI Conference on Artificial Intelligence}, 2022.

\bibitem[Potdar et~al.(2005)Potdar, Han, and Chang]{potdar2005survey}
V.~M. Potdar, S.~Han, and E.~Chang.
\newblock A survey of digital image watermarking techniques.
\newblock In \emph{INDIN'05. 2005 3rd IEEE International Conference on Industrial Informatics, 2005.}, pages 709--716. IEEE, 2005.

\bibitem[Qin et~al.(2022)Qin, Zhang, Chen, Lakshminarayanan, Beutel, and Wang]{PatchRobust_22}
Y.~Qin, C.~Zhang, T.~Chen, B.~Lakshminarayanan, A.~Beutel, and X.~Wang.
\newblock Understanding and improving robustness of vision transformers through patch-based negative augmentation.
\newblock In S.~Koyejo, S.~Mohamed, A.~Agarwal, D.~Belgrave, K.~Cho, and A.~Oh, editors, \emph{NeurIPS}, pages 16276--16289, 2022.

\bibitem[Rebuffi et~al.(2021)Rebuffi, Gowal, Calian, Stimberg, Wiles, and Mann]{rebuffi_aug}
S.-A. Rebuffi, S.~Gowal, D.~A. Calian, F.~Stimberg, O.~Wiles, and T.~A. Mann.
\newblock Data augmentation can improve robustness.
\newblock In M.~Ranzato, A.~Beygelzimer, Y.~Dauphin, P.~Liang, and J.~W. Vaughan, editors, \emph{NeurIPS}, volume~34, pages 29935--29948. Curran Associates, Inc., 2021.
\newblock URL \url{https://proceedings.neurips.cc/paper_files/paper/2021/file/fb4c48608ce8825b558ccf07169a3421-Paper.pdf}.

\bibitem[Recht et~al.(2019)Recht, Roelofs, Schmidt, and Shankar]{recht2019imagenet}
B.~Recht, R.~Roelofs, L.~Schmidt, and V.~Shankar.
\newblock Do imagenet classifiers generalize to imagenet?
\newblock In \emph{International Conference on Machine Learning}, pages 5389--5400. PMLR, 2019.

\bibitem[Rice et~al.(2020)Rice, Wong, and Kolter]{pmlr-v119-rice20a}
L.~Rice, E.~Wong, and Z.~Kolter.
\newblock Overfitting in adversarially robust deep learning.
\newblock In H.~D. III and A.~Singh, editors, \emph{Proceedings of the 37th International Conference on Machine Learning}, volume 119 of \emph{Proceedings of Machine Learning Research}, pages 8093--8104. PMLR, 13--18 Jul 2020.
\newblock URL \url{https://proceedings.mlr.press/v119/rice20a.html}.

\bibitem[Selvaraju et~al.(2017)Selvaraju, Cogswell, Das, Vedantam, Parikh, and Batra]{gradcam}
R.~R. Selvaraju, M.~Cogswell, A.~Das, R.~Vedantam, D.~Parikh, and D.~Batra.
\newblock Grad-cam: Visual explanations from deep networks via gradient-based localization.
\newblock In \emph{ICCV}, pages 618--626. IEEE Computer Society, 2017.
\newblock ISBN 978-1-5386-1032-9.
\newblock URL \url{http://dblp.uni-trier.de/db/conf/iccv/iccv2017.html#SelvarajuCDVPB17}.

\bibitem[Shao et~al.(2022)Shao, Montasser, and Blum]{shao2022theory}
H.~Shao, O.~Montasser, and A.~Blum.
\newblock A theory of pac learnability under transformation invariances.
\newblock \emph{Advances in Neural Information Processing Systems}, 35:\penalty0 13989--14001, 2022.

\bibitem[Shao et~al.(2021)Shao, Shi, Yi, Chen, and Hsieh]{shao2021adversarial}
R.~Shao, Z.~Shi, J.~Yi, P.-Y. Chen, and C.-J. Hsieh.
\newblock On the adversarial robustness of vision transformers.
\newblock \emph{arXiv preprint arXiv:2103.15670}, 2021.

\bibitem[Sonoda et~al.(2021)Sonoda, Ishikawa, and Ikeda]{ghosts}
S.~Sonoda, I.~Ishikawa, and M.~Ikeda.
\newblock Ghosts in neural networks: Existence, structure and role of infinite-dimensional null space.
\newblock \emph{CoRR}, abs/2106.04770, 2021.
\newblock URL \url{https://arxiv.org/abs/2106.04770}.

\bibitem[Steiner et~al.(2022)Steiner, Kolesnikov, Zhai, Wightman, Uszkoreit, and Beyer]{steiner2022how}
A.~P. Steiner, A.~Kolesnikov, X.~Zhai, R.~Wightman, J.~Uszkoreit, and L.~Beyer.
\newblock How to train your vit? data, augmentation, and regularization in vision transformers.
\newblock \emph{Transactions on Machine Learning Research}, 2022.
\newblock ISSN 2835-8856.
\newblock URL \url{https://openreview.net/forum?id=4nPswr1KcP}.

\bibitem[Strang(2009{\natexlab{a}})]{Strang2009Introduction}
G.~Strang.
\newblock \emph{Introduction to Linear Algebra, Fourth Edition}.
\newblock Wellesley Cambridge Press, Feb. 2009{\natexlab{a}}.
\newblock ISBN 0980232716.
\newblock URL \url{http://www.amazon.com/exec/obidos/redirect?tag=citeulike07-20\&path=ASIN/0980232716}.

\bibitem[Strang(2009{\natexlab{b}})]{strang09}
G.~Strang.
\newblock \emph{Introduction to Linear Algebra}.
\newblock Wellesley-Cambridge Press, Wellesley, MA, fourth edition, 2009{\natexlab{b}}.
\newblock ISBN 9780980232714 0980232716 9780980232721 0980232724 9788175968110 8175968117.

\bibitem[Wang et~al.(2019)Wang, Ge, Lipton, and Xing]{wang2019learning}
H.~Wang, S.~Ge, Z.~Lipton, and E.~P. Xing.
\newblock Learning robust global representations by penalizing local predictive power.
\newblock In \emph{Advances in Neural Information Processing Systems}, pages 10506--10518, 2019.

\bibitem[Wang et~al.(2022)Wang, Lan, Liu, Ouyang, Qin, Lu, Chen, Zeng, and Yu]{wang2022generalizing}
J.~Wang, C.~Lan, C.~Liu, Y.~Ouyang, T.~Qin, W.~Lu, Y.~Chen, W.~Zeng, and P.~Yu.
\newblock Generalizing to unseen domains: A survey on domain generalization.
\newblock \emph{IEEE Transactions on Knowledge and Data Engineering}, 2022.

\bibitem[Wang et~al.(2021)Wang, Li, Sun, and Xu]{Wang_2021_CVPR}
S.~Wang, X.~Li, J.~Sun, and Z.~Xu.
\newblock Training networks in null space of feature covariance for continual learning.
\newblock In \emph{Proceedings of the IEEE/CVF Conference on Computer Vision and Pattern Recognition (CVPR)}, pages 184--193, June 2021.

\bibitem[Wolfgang and Delp(1996)]{Wolfgang1996AWF}
R.~B. Wolfgang and E.~J. Delp.
\newblock A watermark for digital images.
\newblock \emph{Proceedings of 3rd IEEE International Conference on Image Processing}, 3:\penalty0 219--222 vol.3, 1996.

\bibitem[Xiao et~al.(2023)Xiao, Tang, Wei, Liu, and Lin]{xiao2023masked}
Y.~Xiao, Z.~Tang, P.~Wei, C.~Liu, and L.~Lin.
\newblock Masked images are counterfactual samples for robust fine-tuning.
\newblock In \emph{Proceedings of the IEEE/CVF Conference on Computer Vision and Pattern Recognition}, pages 20301--20310, 2023.

\bibitem[Yeh et~al.(2023)Yeh, Chen, Wu, Chen, Vi{\'e}gas, and Wattenberg]{yeh2023attentionviz}
C.~Yeh, Y.~Chen, A.~Wu, C.~Chen, F.~Vi{\'e}gas, and M.~Wattenberg.
\newblock Attentionviz: A global view of transformer attention.
\newblock \emph{arXiv preprint arXiv:2305.03210}, 2023.

\bibitem[Yun et~al.(2019)Yun, Han, Oh, Chun, Choe, and Yoo]{yun2019cutmix}
S.~Yun, D.~Han, S.~J. Oh, S.~Chun, J.~Choe, and Y.~Yoo.
\newblock {CutMix}: Regularization strategy to train strong classifiers with localizable features.
\newblock In \emph{ICCV}, 2019.

\bibitem[Zhang et~al.(2018)Zhang, Cisse, Dauphin, and Lopez-Paz]{zhang2018mixup}
H.~Zhang, M.~Cisse, Y.~N. Dauphin, and D.~Lopez-Paz.
\newblock mixup: Beyond empirical risk minimization.
\newblock In \emph{ICLR}, 2018.
\newblock URL \url{https://openreview.net/forum?id=r1Ddp1-Rb}.

\bibitem[Zhang et~al.(2019)Zhang, Yu, Jiao, Xing, El~Ghaoui, and Jordan]{trades}
H.~Zhang, Y.~Yu, J.~Jiao, E.~Xing, L.~El~Ghaoui, and M.~Jordan.
\newblock Theoretically principled trade-off between robustness and accuracy.
\newblock In \emph{ICML}, pages 7472--7482. PMLR, 2019.

\bibitem[Zou et~al.(2023)Zou, Yang, Zhang, Li, Li, Gao, and Lee]{zou2023segment}
X.~Zou, J.~Yang, H.~Zhang, F.~Li, L.~Li, J.~Gao, and Y.~J. Lee.
\newblock Segment everything everywhere all at once, 2023.

\end{thebibliography}

\newpage
\onecolumn
{
\centering
\section*{Supplementary Material}
}
\appendix

\label{sec:appendix}
\section{Proof of Proposition 1}

Let $d$ be the hidden dimension of the attention layer. $\mathbf{Q}_i, \mathbf{K}_i \in \mathbb{R}^{d \times d_k}$ where $d_k = d/h$. $\operatorname{rank}(\mathbf{Q}_i\mathbf{K}_i^\top) \leq \operatorname{rank}(\mathbf{K}_i^\top) \leq d_k$. 
Consider the sum of row spaces $S = \operatorname{R}(\mathbf{Q}_1\mathbf{K}_1^\top) + \operatorname{R}(\mathbf{Q}_2\mathbf{K}_2^\top) + \cdots + \operatorname{R}(\mathbf{Q}_h\mathbf{K}_h^\top)$. $S$ is a subspace of $\mathbb{R}^d$. For $i = 1, \dots, h$, choose a basis for $\operatorname{R}(\mathbf{Q}_i\mathbf{K}_i^\top)$, denoted as $B_i = \{\mathbf{b}_1, \cdots, \mathbf{b}_{n_i}\}$, $|B_i| = n_i \leq d_k $. Without loss of generality, let $\mathbf{r}_{m, k} \in B_m$. 

$S = \operatorname{span}(\bigcup\limits_{i=1}^h B_i)$, so 

\begin{equation}
\begin{split}
\operatorname{dim}\left(S\right) &= \operatorname{dim}\left(\operatorname{span}\left(\bigcup\limits_{i=1}^h B_i\right)\right) 
= \operatorname{dim}\left(\operatorname{span}\left(\left( \bigcup\limits_{\substack{i=1 \\ i\neq m}}^h B_i \right) \cup \left(B_m \setminus \{\mathbf{r}_{m, k} \} \right)\right)\right) \\
&\leq \left\lvert \left( \bigcup\limits_{\substack{i=1 \\ i\neq m}}^h B_i \right) \cup \left(B_m \setminus \{\mathbf{r}_{m, k} \} \right) \right\rvert
\leq (h - 1)d_k + (d_k - 1) = d - 1.
\end{split}
\end{equation}

So, $\exists \mathbf{w} \in \mathbb{R}^d, \mathbf{w} \neq \mathbf{0}$ and $\mathbf{w} \in S^\perp$. This means for $i = 1, \dots, h, \mathbf{w} \in \left(\operatorname{R}\left(\mathbf{Q}_i\mathbf{K}_i^\top\right)\right)^\perp, \mathbf{w} \in \operatorname{N}\left(\mathbf{Q}_i\mathbf{K}_i^\top\right).$ By condition 2, $\operatorname{N}(\mathbf{V}_i) \supseteq \operatorname{N}(\mathbf{Q}_i\mathbf{K}_i^\top)$, so $\mathbf{w} \in \operatorname{N}\left(\mathbf{Q}_i\mathbf{K}_i^\top\right) \cap \operatorname{N}(\mathbf{V}_i^\top).$

Then, we can choose $\mathbf{W}$ wherein each row is a multiple of $\mathbf{w}$. We have $\mathbf{WV}_i = \mathbf{0}$, and for any input to the encoder $\mathbf{X} \in \mathbb{R}^{n \times d}$,
\begin{equation}
    \mathbf{WQ}_i\mathbf{K}_i^\top\mathbf{X}^\top + \mathbf{XQ}_i\mathbf{K}_i^\top\mathbf{W}^\top + \mathbf{WQ}_i\mathbf{K}_i^\top\mathbf{W}^\top = \mathbf{0}.
\end{equation}

Consider the output of attention head,
\begin{equation}
\begin{split}
\operatorname{head}_i(\mathbf{X+W}) &= \operatorname{Softmax}\left(\frac{\mathbf{\left(X+W\right)Q}_i\mathbf{K}_i^\top\mathbf{\left(X+W\right)}^\top}{\sqrt{d_k}}\right)\mathbf{(X+W)V}_i \\
&= \operatorname{Softmax}\left(\frac{\mathbf{XQ}_i\mathbf{K}_i^\top\mathbf{X}^\top + \mathbf{WQ}_i\mathbf{K}_i^\top\mathbf{X}^\top + \mathbf{XQ}_i\mathbf{K}_i^\top\mathbf{W}^\top + \mathbf{WQ}_i\mathbf{K}_i^\top\mathbf{W}^\top}{\sqrt{d_k}}\right)\mathbf{XV}_i\\
&= \operatorname{Softmax}\left(\frac{\mathbf{XQ}_i\mathbf{K}_i^\top\mathbf{X}^\top}{\sqrt{d_k}}\right)\mathbf{XV}_i = \operatorname{head}_i(\mathbf{X}).
\end{split}
\end{equation}
Adding the noise $\mathbf{W}$ does not change the output of any attention head for arbitrary input $\mathbf{X}$, which completes our proof. 

\newpage
\section{Algorithm and implementation details} \label{sec:app_alg}

We present the algorithm of our data augmentation with nullspace noise in \cref{alg_ns}. 

\SetKw{Break}{break}
\IncMargin{1.2em}
\begin{algorithm}[H]
\caption{Data augmentation with nullspace noise}\label{alg_ns}
\textbf{Input: } transformer model with patch embedding layer $f_e$, encoder $f_\phi$ and linear classifier $f_\psi$ parameterized by $e, \phi, \psi$ respectively; training data $\mathcal{T}$; batch size $B$; sampling limit $A$; noise nullity threshold $\epsilon$; noise learning rate $\eta_{v}$; model learning rate $\eta_{f}$; number of outer iterations $K$, noise training step $T$, model training step $S$

\For {$k = 0, \cdots, K-1$}{
    Sample initial noise $\mathbf{v} \sim \operatorname{U}(-\textrm{lim}, \textrm{lim})$ \;
    \For {$t = 0, \cdots, T-1$}{
        Sample a minibatch $(\mathbf{X, y}) \sim \mathcal{T}$ \;
        Compute $\mathbf{U} \gets f_e(\mathbf{X})$ \;
        Compute logits $\mathbf{Z} \gets f_\psi(f_\phi^{0}(\mathbf{U}))$, $\mathbf{Z}^\prime \gets f_\psi(f_\phi^{0}(\mathbf{U} + [\mathbf{v}]))$  \quad \texttt{\# "[v]" means broadcasting the noise v along the sample dimension}\;
        Compute $\delta \gets \frac{1}{B}\sum_{i=1}^B \lVert \operatorname{Softmax}(\mathbf{z}_i^\prime) - \operatorname{Softmax}(\mathbf{z}_i) \rVert^2$ \quad \texttt{\#} $\mathbf{z}_i$ \texttt{is sample logit}\;
        \If{$ \sigma < \epsilon$}
        {\Break \;}
        Calculate $\ell \gets \frac{1}{B}\sum_{i=1}^B \lVert \mathbf{z}_i^\prime - \mathbf{z}_i \rVert^2$ \;
        Update $\mathbf{v} \gets \mathbf{v} - \nabla_{\mathbf{v}} \ell$
        
    }
    \For {$s = 0, \cdots, S-1$}{
    Sample a minibatch $(\mathbf{X, y}) \sim \mathcal{T}$ \;
    Compute $\mathbf{U} \gets f_e(\mathbf{X})$ \;
    Compute logits $\mathbf{Z} \gets f_\psi(f_\phi^{0}(\mathbf{U}))$, $\mathbf{Z}^\prime \gets f_\psi(f_\phi^{0}(\mathbf{U} + [\mathbf{v}]))$ \;
    Compute loss $\mathcal{L} \gets \frac{1}{B} \sum_{i=1}^B (\ell(\mathbf{z}_i, y_i) + \ell(\mathbf{z}_i^\prime, y_i))$, where $\ell$ is the cross-entropy loss \;
    Update model parameters $(\psi, \phi, e) \gets (\psi, \phi, e) - \nabla_{(\psi, \phi, e)} \mathcal{L}$\;
    }
}
\textbf{Output: model weight $(\psi, \phi, e)$}
\end{algorithm}
\DecMargin{1.2em}


\paragraph{Hyperparameters} We fine-tuned the ViT model for $K=20$ rounds in all settings. In each round, we initialized the noise with sampling limit $A=3$, optimized it with learning rate $\eta_v = 0.1$ and set a threshold of $\epsilon = 0.03$. We empirically found that $T=3000$ is enough to trigger early stopping so that the learned noise satisfies the $\epsilon$ threshold. We used $\eta_f = 10^{-5}$ to fine-tune the model for $S=40$ iterations in each round. We set batch size $B=128$, and slightly different from the vanilla SGD in Alg~\ref{alg_ns}, we used the AdamW optimizer ~\citep{loshchilov2018decoupled} and cosine learning rate scheduler with defualt hyperparameters for both the noise and the model training.

The original \texttt{ViT-B + DAT} model~\citep{mao2022enhance} used the Exponential Moving Average (EMA) for evaluation\footnote{\url{https://github.com/alibaba/easyrobust}}, so we also used EMA to evaluate the performance of \texttt{ViT-B + DAT} fine-tuned with our method. For all the other settings, we used single model without ensemble for evaluation. We used $\epsilon = 1/255$ for the FGSM attack consistent with ~\cite{mao2022enhance}. 

\paragraph{Computation time} The experiments were conducted on a combination of A100, V100 GPUs and a 3090 GPU, depending on the availability. Although we only used about 10\% of the ImageNet-1k~\citep{deng2009imagenet} training data to fine-tune the model, the main computation time is on training the nullspace noise. One run of our experiment (20 rounds) takes the time roughly equivalent to 8 epochs of standard training on ImageNet-1k.

\section{Change trend of the noise influence with the fine-tuning steps}
\label{sec:validate_mse}
Beside the trend of noise norm and performance metrics in Fig.~\ref{fig:trend}, we also keep track of the influence of the learned noise in terms of MSE probability (\ref{sec:vits:2}) at every 80 steps of the model fine-tuning. As shown in Table~\ref{tab:confirm_mse}, the noise influence is always lower than $\epsilon=0.03$, which means early stopping is triggered and the model enters the $\epsilon$ region.

\begin{table}[!ht]
\centering
\caption{MSE probability of the noise at different fine-tuning steps.}
\resizebox{\textwidth}{!}{%
\begin{tabular}{ccccccccccc}
\toprule
\textbf{Fine-Tuning Step} & 40 & 120 & 160 & 280 & 360 & 440 & 520 & 600 & 680 & 760 \\ 
\midrule
\textbf{MSE Probability} & 0.028 & 0.027 & 0.026 & 0.029 & 0.028 & 0.029 & 0.027 & 0.025 & 0.028 & 0.026 \\ 
\bottomrule
\end{tabular}
}
\label{tab:confirm_mse}
\end{table}

\begin{figure}[h!]
  \centering
  \begin{subfigure}[c]{0.46\columnwidth}
    \centering
    \includegraphics[width=\linewidth]{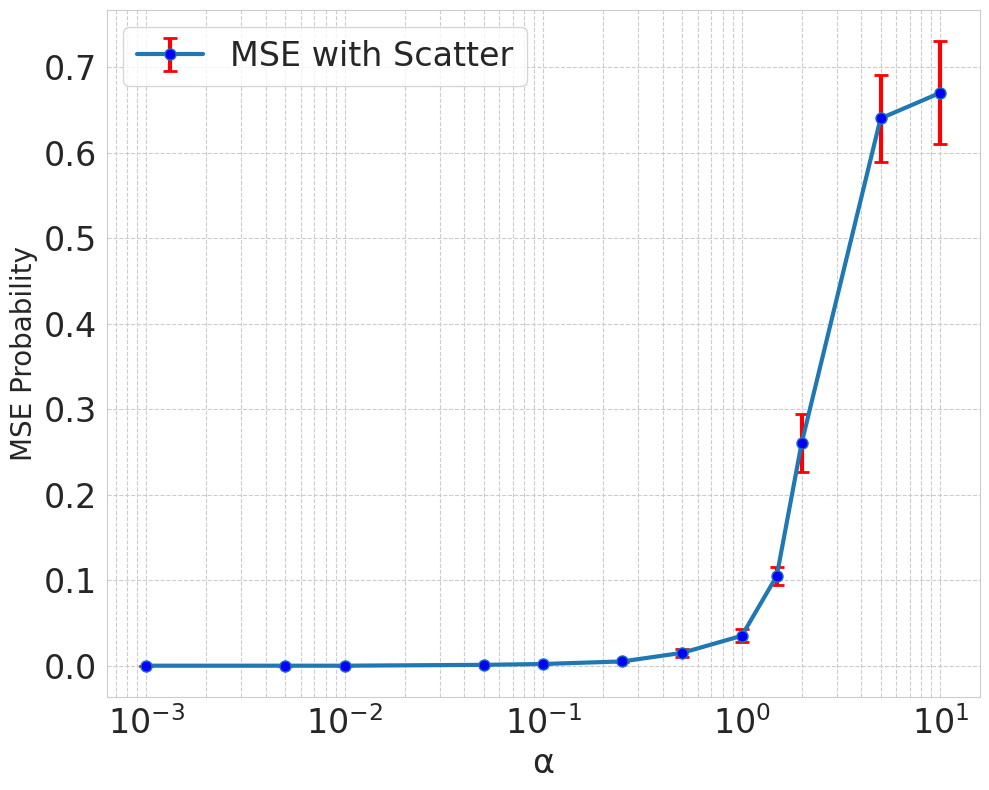}
    \caption[Short caption]{\scriptsize{Influence of $\epsilon$ noise under multiplication with different $\alpha$}}
    \label{fig:subfig1}
  \end{subfigure}
  \hfill 
  \begin{subfigure}[c]{0.46\columnwidth}
    \centering
    \includegraphics[width=0.95\linewidth]{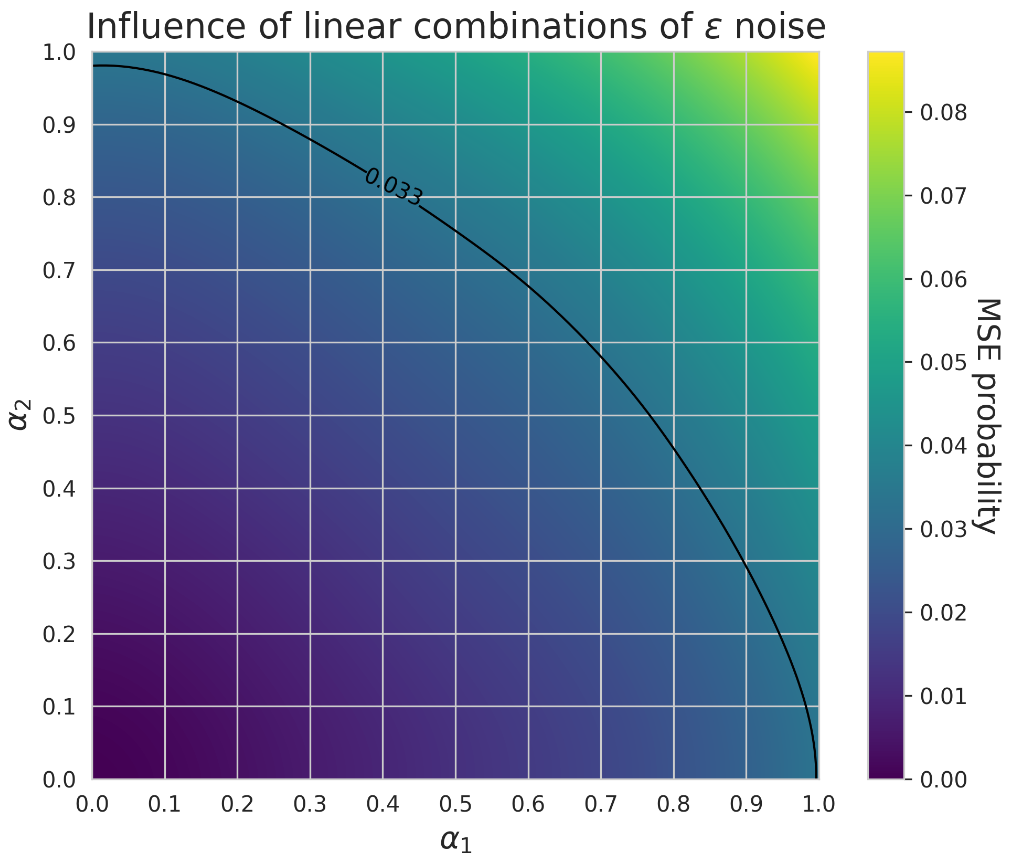}
    \caption[Short caption]{\scriptsize{Influence of $\epsilon$ noise under convex combination with  different $\alpha_1, \alpha_2$}}
    \label{fig:subfig2}
  \end{subfigure}
  \caption{Validation of the properties of the $\epsilon$-approximate nullspace.}
  \label{fig:approx_ns_axioms}
\end{figure}

\section{Approximate Nullspace Properties}
To explore the property of the $\epsilon$-approximate nullspace, we conduct an experiment to observe the behavior of the learned noise vectors under scalar multiplication and convex combination. For this, we first construct a set of $m$ $\epsilon$-approximate nullspace vectors $\mathbf{V} = {\{\mathbf{v}_i\}}_{i=1}^{m}$ starting from different random initializations using $\epsilon=0.033$. For scalar multiplication, we vary the scaling factor $\alpha$ and report the mean influence of $\alpha \mathbf{v}$ on the model's predictions in terms of MSE probabilities (\cref{fig:approx_ns_axioms}(a)). For convex combination, we sample $n$ different pairs of nullspace vectors from $\mathbf{V}$, denoted as $\mathbf{P} = {\{(\mathbf{v}_{\mathcal{J}_{k, 1}}, \mathbf{v}_{\mathcal{J}_{k, 2}})\}}_{k=1}^{n}$, where $\mathcal{J}_{k, 1}, \mathcal{J}_{k, 2} \in \{1, 2, \ldots, m\}, \forall k \in \{1, \ldots n\}$. Then, we vary $\alpha_1$ and $\alpha_2$ between $[0, 1]$ with a grid size of $0.1$, and for each combination of $(\alpha_1, \alpha_2)$, we evaluate the influence of the convex combination $\alpha_1 \mathbf{v}_{\mathcal{J}_{k, 1}} + \alpha_2 \mathbf{v}_{\mathcal{J}_{k, 2}}$ on the model's prediction in MSE probability, averaged over all values of $k$. In practice we set $m=100, n=10$. The influence of the linear combined noise at each point of the grid is visualized as a heatmap as shown in \cref{fig:approx_ns_axioms}(b).

The results in \cref{fig:approx_ns_axioms} show that the approximate nullspace has similar property to vector space in terms of closure under addition and scalar multiplication within a certain range of coefficients. When the scaling factor $\alpha < 1$, we see a clear trend that the MSE probability of the scaled noise is less than $\alpha \epsilon$. In the linear combination case, the line $\alpha_1 + \alpha_2 = 1$ is well within the contour line of MSE probability being 0.033, showing that the convex combination of a pair of $\epsilon$ noise vector is still $\epsilon$-approximate.

\section{Watermarking Images}
Watermarking as image, usually used to 
convey ownership information or verify content of the data, 
has been studied extensively \citep{Wolfgang1996AWF, potdar2005survey, al2007combined, bhat2010adaptive}. 
A watermark can be either imperceptible or perceptible. 
and perceptible watermarking applies a noticeable marker to convey the protected nature of the data \citep{berghel1998digital}. In this section, we explore to utilize nullspace noise to apply a perceptible watermark 
on the image which does not alter the test-time forward process.

\begin{figure}[!htbp]
    \centering
        \begin{subfigure}{0.45\textwidth}
        \includegraphics[width=1\textwidth]{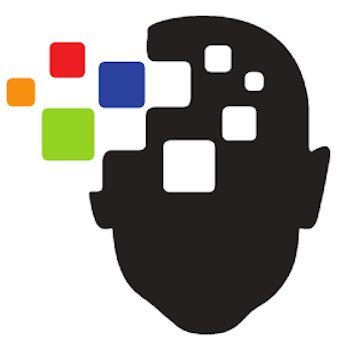}
        \end{subfigure}
        \begin{subfigure}{0.45\textwidth}
        \centering
        \includegraphics[width=1\textwidth]{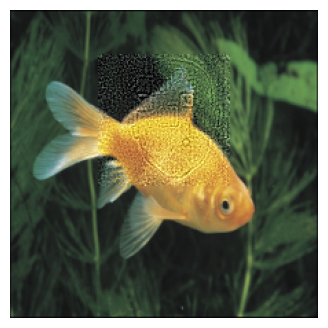}
        \end{subfigure}
        \begin{subfigure}{0.45\textwidth}
        \includegraphics[width=1\textwidth]{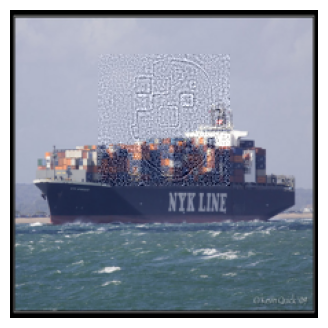}
        \end{subfigure}
        \begin{subfigure}{0.45\textwidth}
        \includegraphics[width=1\textwidth]{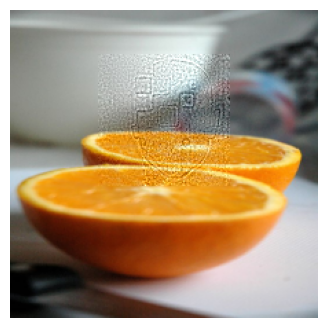}
        \end{subfigure}
        \begin{subfigure}{0.45\textwidth}
        \centering
        \includegraphics[width=1\textwidth]{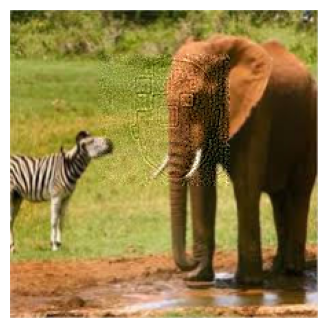}
        \end{subfigure}
        \begin{subfigure}{0.45\textwidth}
        \centering
        \includegraphics[width=1\textwidth]{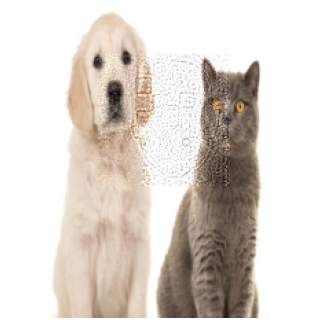}
        \end{subfigure}
        
    \caption{\footnotesize{\textbf{Watermark superposition using the nullspace basis vectors.}}}
    \label{fig:watermarks}
\end{figure}

\cref{fig:watermarks} shows an example watermarking approach using the nullspace noise. Here, we emboss the ICML logo onto the natural images. The resulting modified image, attains the final predictions close to the original image. ($100\%$ match in the final output prediction and $10^{-4}$ difference in the predicted confidence value of the assigned class.)

\textbf{Method details:} With basis vectors of the nullspace, we can construct a watermark to be overlaid on the original image without affecting the output of the network. Given a source (user's image) and a target image (watermark), we simply need to estimate the scalar parameters corresponding to the basis vectors to satisfy $\sum_{i=0}^{i<m} \mathbf{e}_i\lambda_i =  \mathbf{v}_\theta \approx \Delta \mathbf{x}_j$.   

$\mathbf{e}_i$ are the basis vectors for the nullspace, $\lambda_i$ are their corresponding scalar co-efficients which are to be determined and $\Delta \mathbf{x}_j$ is the changed required to convert $j^{\text{th}}$ original image patch to $j^{\text{th}}$ watermark image patch. This can be achieved through a constrained optimisation of the following form:

\begin{align}
    \min \lVert \Delta \mathbf{x}_j - \sum_{i=0}^{i<m} \mathbf{e}_i\lambda_i\rVert_p.
    \label{eq:min_watermark}
\end{align}

Here, $\Delta x_j$ is the difference between the $j^{th}$ channel of a source and target image and $\lambda_i$ is the $i^{th}$ nullspace basis vector of the patch embedding layer with the corresponding variable scalar $e_i$. We use a least-square solver to solve for the solution (Available readily with packages such as Numpy).

\section{Targeted Nullspace Noise}

\begin{figure}[!htbp]
    \centering
    \footnotesize
    \begin{subfigure}{1\textwidth}
        \begin{subfigure}{0.49\textwidth}
        \begin{framed}
            \centering
            \includegraphics[width=1\textwidth]{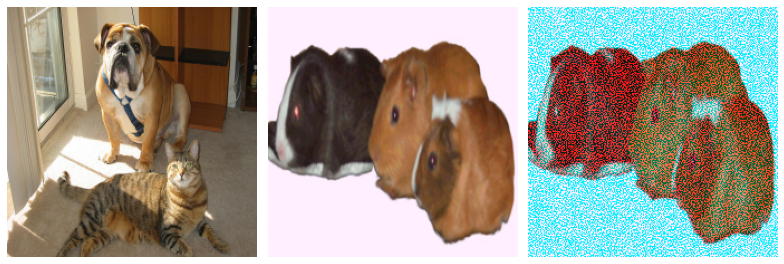}
        \end{framed}
        \end{subfigure}
        \begin{subfigure}{0.49\textwidth}
            \begin{framed}
            \centering
            \includegraphics[width=1\textwidth]{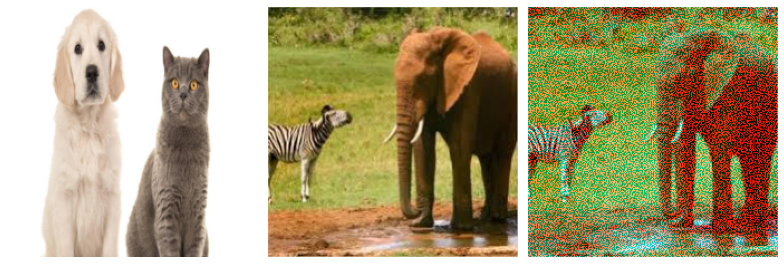}
            \end{framed}
        \end{subfigure}
        \caption{Triplet of Source, target and transformed images}
    \end{subfigure}
    \begin{subfigure}{1\textwidth}
            \begin{subfigure}{0.49\textwidth}
            \centering
            \includegraphics[width=1\textwidth]{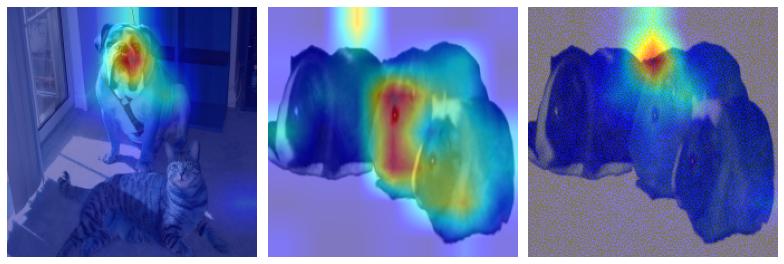}
        \end{subfigure}
        \begin{subfigure}{0.49\textwidth}
            \centering
            \includegraphics[width=1\textwidth]{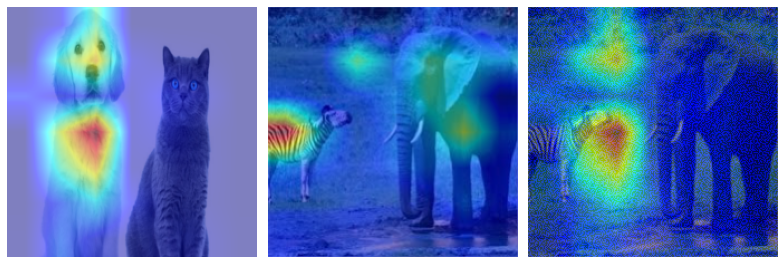}
        \end{subfigure}
        \caption{Saliency maps for the corresponding images from the row above.}
    \end{subfigure}
    
    \caption{{\textbf{Targeted nullspace noise.} Transformed images appear visually as target images but are interpreted as source images by the model. The equivalence between source and transformed images is not only in terms of the final predictions but also in the interpretability maps depicted in (b).}}
    \label{fig:targets}
\end{figure}

Due to the dimension reduction effect of the patch embedding layer in most ViTs, we can transfer an image to be visually similar to another image by human perception, without changing the output of the original image perceived by the model. This differs from adversarial examples in the following aspects:
\begin{enumerate}
    \item The working direction to construct an adversarial example is the other way around. If the transformed image is to be considered an adversarial example, then our source becomes the target for adversarial training and our target becomes the source. 
    \item Generating targeted nullspace noise requires no backpropagation through the network
    \item Not only does the final prediction on the transformed image matches the source image, the saliency maps also match. This is displayed in Fig. \ref{fig:targets}
\end{enumerate}

Though the transformation is not perfect, we can spot that the transformed images are visually similar to target images rather than source images. 
Even with this difference in the input space, transformed images and source images are classified into the same category with roughly the same confidence.  

As recent studies have shown, fooling can also be extended 
to the interpretability methods (XAI) \citet{dombrowski2019explanations} partially due the limitations exposed by recent studies \citep{dombrowski2019explanations, ghorbani2019interpretation, heo2019fooling}. 
However, in contrast to these works aiming to fool specific XAI method, our nullspace noise only depends on the model, not the XAI method. 

In Fig. \ref{fig:targets}(b), we show the interpretability maps as generated by LRP \citep{Chefer_2021_CVPR}. From the figure, we can observe that the heatmaps generated by source and transformed images are identical whereas, the transformed image heatmaps substantially differ from target images'. Though only reported for LRP, we observed that a similar observation holds across different interpretability approaches. Here, we only presented the results on LRP, as in the context of ViTs, we found the heatmaps from other methods to be lacking (also pointed out by authors of LRP).

\begin{figure}[t!]
    \centering
    \begin{subfigure}{0.49\textwidth}
        \FrameSep2pt
        \begin{framed}
        \begin{subfigure}{1.0\textwidth}
            \includegraphics[width=1\textwidth]{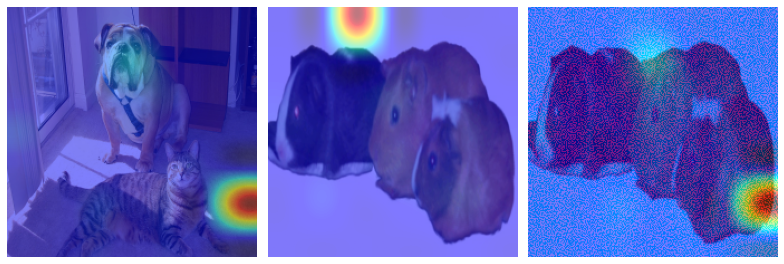}
        \end{subfigure}
        \begin{subfigure}{1.0\textwidth}
            \includegraphics[width=1\textwidth]{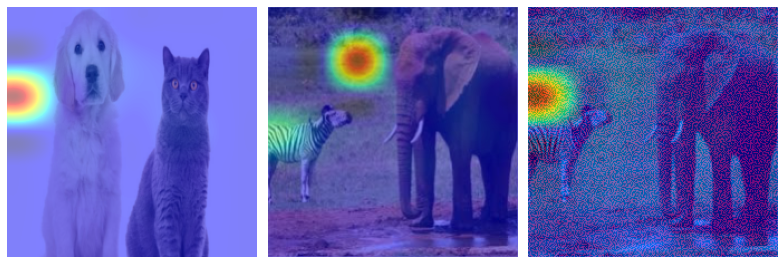}
        \end{subfigure}
        \begin{subfigure}{1.0\textwidth}
            \includegraphics[width=1\textwidth]{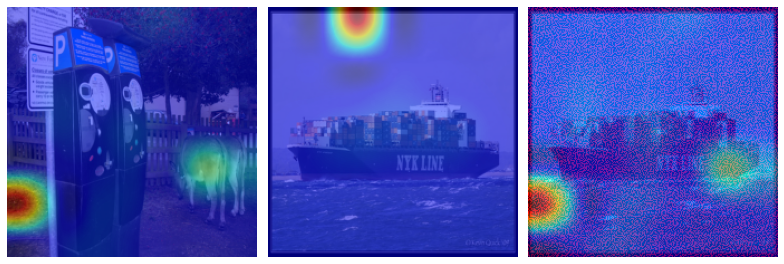}
        \end{subfigure}
        \end{framed}
        \caption{Attention \citep{Chefer_2021_CVPR}}
    \end{subfigure}
    \begin{subfigure}{0.49\textwidth}
        \FrameSep2pt
        \begin{framed}
        \begin{subfigure}{1.0\textwidth}
            \includegraphics[width=1\textwidth]{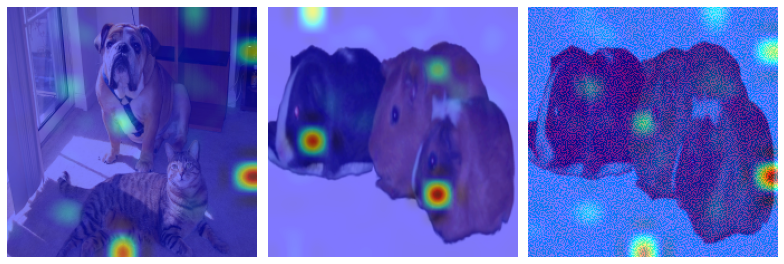}
        \end{subfigure}
        \begin{subfigure}{1.0\textwidth}
            \includegraphics[width=1\textwidth]{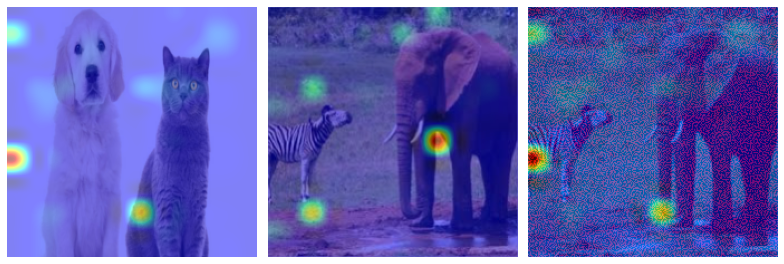}
        \end{subfigure}
        \begin{subfigure}{1.0\textwidth}
            \includegraphics[width=1\textwidth]{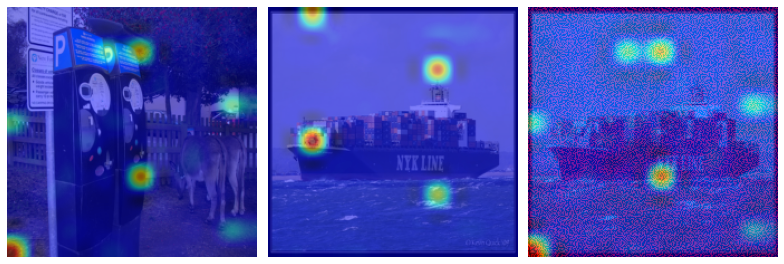}
        \end{subfigure}
        \end{framed}
    \caption{Grad-CAM \citep{gradcam}}
    \end{subfigure}
    \begin{subfigure}{0.49\textwidth}
        \FrameSep2pt
        \begin{framed}
        \begin{subfigure}{1.0\textwidth}
            \includegraphics[width=1\textwidth]{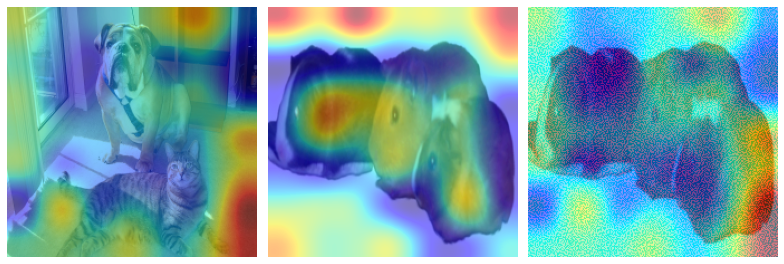}
        \end{subfigure}
        \begin{subfigure}{1.0\textwidth}
            \includegraphics[width=1\textwidth]{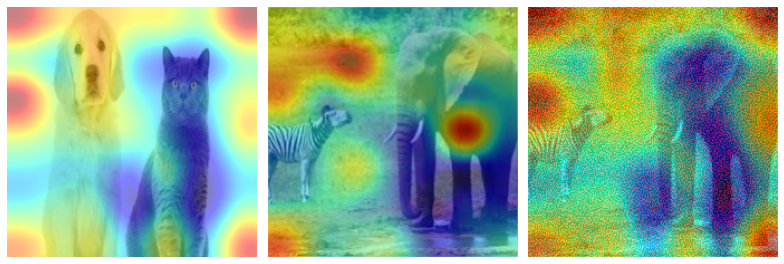}
        \end{subfigure}
        \begin{subfigure}{1.0\textwidth}
            \includegraphics[width=1\textwidth]{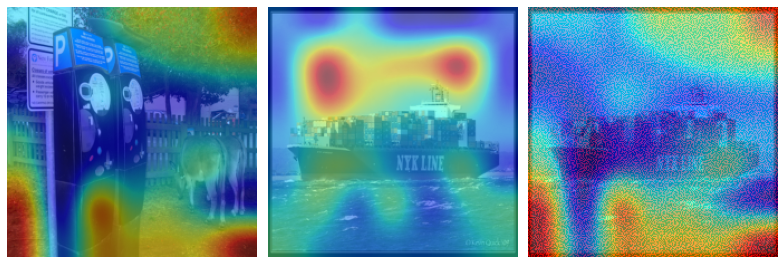}
        \end{subfigure}
        \end{framed}
        \caption{Rollout \citep{rollout}}
    \end{subfigure}
        \begin{subfigure}{0.49\textwidth}
        \FrameSep2pt
        \begin{framed}
        \begin{subfigure}{1.0\textwidth}
            \includegraphics[width=1\textwidth]{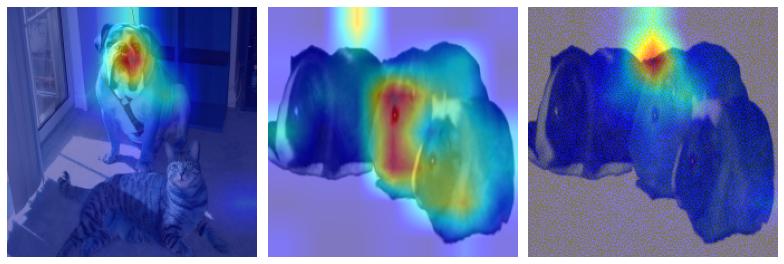}
        \end{subfigure}
        \begin{subfigure}{1.0\textwidth}
            \includegraphics[width=1\textwidth]{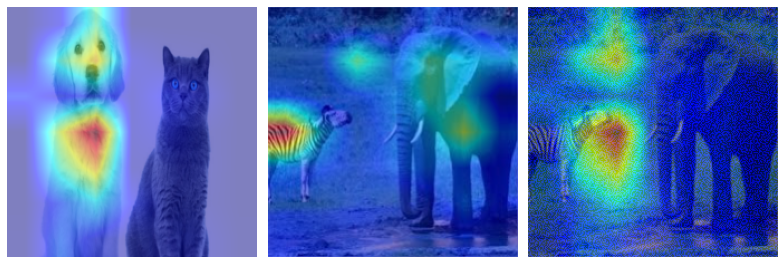}
        \end{subfigure}
        \begin{subfigure}{1.0\textwidth}
            \includegraphics[width=1\textwidth]{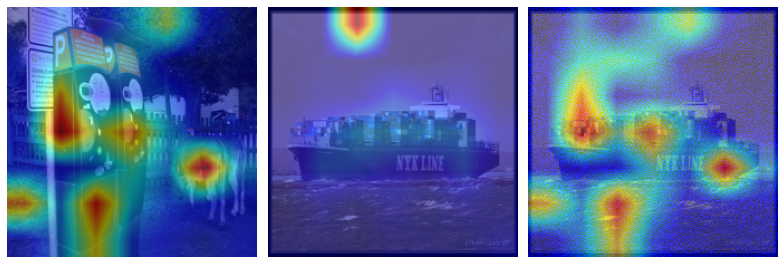}
        \end{subfigure}
        \end{framed}
        \caption{LRP \citep{Chefer_2021_CVPR}}
    \end{subfigure}
    \caption{\textbf{Interpretability maps generated via different methods for (source, target, transformed) images}}
    \label{fig:app_im_xai}
\end{figure}

In Fig. \ref{fig:app_im_xai} we show the saliency maps generated by different XAI methods. Even though the maps generated by methods other than LRP are poor (hard to interpret), we see that the source and transformed respond similarly to these methods.

\end{document}